\documentclass[5p]{elsarticle}

\usepackage{algorithm}
\usepackage{algpseudocode}
\usepackage{amsmath}
\usepackage{amssymb}
\usepackage{graphicx}
\usepackage{lettrine}
\usepackage{rotating}
\usepackage{threeparttable}
\usepackage[table]{xcolor}

\begin{document}

\begin{frontmatter}

\title{A Social Spider Algorithm for Global Optimization}

\author[hku]{James J.Q. Yu\corref{cor}}
\ead{jqyu@eee.hku.hk}
\author[hku]{Victor O.K. Li}
\ead{vli@eee.hku.hk}
\cortext[cor]{Corresponding author}
\address[hku]{Department of Electrical and Electronic Engineering, The University of Hong Kong, Pokfulam, Hong Kong}

\begin{abstract}
The growing complexity of real-world problems has motivated computer scientists to search for efficient problem-solving methods. Metaheuristics based on evolutionary computation and swarm intelligence are outstanding examples of nature-inspired solution techniques. Inspired by the social spiders, we propose a novel Social Spider Algorithm to solve global optimization problems. This algorithm is mainly based on the foraging strategy of social spiders, utilizing the vibrations on the spider web to determine the positions of preys. Different from the previously proposed swarm intelligence algorithms, we introduce a new social animal foraging strategy model to solve optimization problems. In addition, we perform preliminary parameter sensitivity analysis for our proposed algorithm, developing guidelines for choosing the parameter values. The Social Spider Algorithm is evaluated by a series of widely-used benchmark functions, and our proposed algorithm has superior performance compared with other state-of-the-art metaheuristics. 
\end{abstract}


\end{frontmatter}

\section{Introduction}\label{sec:introduction}

\lettrine[lines=2]{W}{ith} the fast growing size and complexity of modern optimization problems, evolutionary computing is becoming increasingly attractive as an efficient tool for optimization. Depending on the nature of phenomenon simulated, evolutionary computing algorithms can be classified into two important groups: evolutionary algorithms (EAs) and swarm intelligence based algorithms. EAs, which mainly draw inspiration from nature, have been shown to be very successful for optimization among all the methods devised by the evolutionary computation community. Currently several types of EAs have been widely employed to solve real world combinatorial or global optimization problems, including Genetic Algorithm (GA), Genetic programming (GP), Evolutionary Strategy (ES) and Differential Evolution (DE). These algorithms demonstrate satisfactory performance compared with conventional optimization techniques, especially when applied to solve non-convex optimization problems \cite{Talbi2009Metaheuristics:FromDesign}\cite{MallipeddiMallipeddiSuganthanTasgetiren2011DifferentialEvolutionAlgorithm}.

In the past two decades, swarm intelligence, a new kind of evolutionary computing technique, has attracted much research interest \cite{ParpinelliLopes2011Newinspirationsin}. The term swarm is employed in a general manner to refer to any collection of interactive agents. Swarm intelligence is mainly concerned with the methodology to model the behavior of social animals and insects for problem solving. Researchers devised optimization algorithms by mimicking the behavior of ants, bees, bacteria, fireflies and other organisms. The impetus of creating such algorithms was provided by the growing needs to solve optimization problems that were very difficult or even considered intractable.

Among the commonly seen animals, spiders have been a major research subject in bionic engineering for many years. However, most research related to spiders focused on the imitation of its walking pattern to design robots, e.g. \cite{YimZhangDuff2002ModularRobots}. A possible reason for this is that a majority of the spiders observed are solitary \cite{Foelix1996BiologySpiders}, which means that they spend most of their lives without interacting with others of their species. However, among the 35 000 spider species observed and described by scientists, some species are social. These spiders, e.g. \textit{Mallos gregalis} and \textit{Oecobius civitas}, live in groups and interact with others in the same group. Based on these social spiders, this paper formulates a new global optimization method to solve optimization problems.

Spiders are air-breathing arthropods. They have eight legs and chelicerae with fangs. Spiders have been found worldwide and are one of the most diverged species among all groups of organisms. They use a wide range of strategies for foraging, and most of them detect prey by sensing vibrations. Spiders have long been known to be very sensitive to vibratory stimulation, as vibrations on their webs notify them of the capture of prey. If the vibrations are in a defined range of frequency, spiders attack the vibration source. The social spiders can also distinguish vibrations generated by the prey with ones generated by other spiders \cite{SchaberGorbBarth2012ForceTransformationIn}. The social spiders passively receive the vibrations generated by other spiders on the same web to have a clear view of the web. This is one of the unique characteristics which distinguishes the social spiders from other organisms as the latter usually exchange information actively, which reduces the information loss to some degree but increases the energy used per communication action \cite{StoddardSalazar2011Energeticcostcommunication}.

The group living phenomenon has been studied intensively in animal behavior ecology. One of the reasons that animals gather and live together is to increase the possibility of successful foraging and reduce the energy cost in this process \cite{HouseLandisUmberson1988Socialrelationshipsand}. In order to facilitate the analysis of social foraging behaviour, researchers proposed two foraging models: information sharing (IS) model \cite{ClarkMangel1984ForagingandFlocking} and producer-scrounger (PS) model \cite{BarnardSibly1981Producersandscroungers:}. The individuals under the IS model perform searching and seek for opportunity to join other individuals simultaneously. In the PS model, the individuals are divided into leaders and followers. Since there is no leader in social spiders \cite{Uetz1992Foragingstrategiesspiders}, the IS model is more suitable to formulate the foraging behavior of social spiders, and we use this model to control the searching pattern of our proposed algorithm.

In this paper, inspired by the social behavior of the social spiders, especially their foraging behavior, we propose a new metaheuristic for global optimization: the Social Spider Algorithm (SSA). The foraging behavior of the social spider can be described as the cooperative movement of the spiders towards the food source position. The spiders receive and analyze the vibrations propagated on the web to determine the potential direction of a food source \cite{Fernandez2007Groupforagingin}. We utilize this natural behavior to perform optimization over the search space in SSA.

The contribution of this paper is threefold:
\begin{itemize}
\item We propose a new nature-inspired swarm intelligence algorithm based on social spiders. This population-based general-purpose metaheuristic demonstrates outstanding performance in the global optimization benchmark tests.
\item We introduce a new social animal foraging model into metaheuristic design. This is the very first attempt of employing the IS model to solve optimization problems. We also incorporate the information loss schemes in the algorithm, which is a unique design of our proposed algorithm.
\item We perform a series of experiments to investigate the impact of different parameters and searching schemes on the performance of the algorithm. The result of these experiments may serve as important inputs for further research.
\end{itemize}

The rest of this paper is organized as follows. We will first present some related work on swarm intelligence and bio-inspired metaheuristics in Section \ref{sec:background}. Then we will formulate and elaborate on SSA by idealizing and imitating the foraging behavior of social spiders in Section \ref{sec:ssa}. Section \ref{sec:benchmark} introduces the benchmark functions we use for testing the performance of SSA, with the experimental settings. Section \ref{sec:result} presents the simulation results of SSA on the benchmark functions and the comparison with other popular metaheuristics. Finally we will conclude this paper in Section \ref{sec:conclusion} and propose some future work.

\section{Background}\label{sec:background}

Swarm intelligence algorithms mimic the methods in nature to drive a search for the optimal solution. At the very beginning there are two major methods for this kind of algorithms: ant colony optimization (ACO) \cite{Dorigo1990OptimizationLearningand} and particle swarm optimization (PSO) \cite{KennedyEberhart1995Particleswarmoptimization}.

ACO is inspired by the foraging behavior of ants, whose goal is to find a shortest path from their colony to food sources. In this metaheuristic, feasible solutions of the optimization problem to be solved are represented by the paths between the colony and food sources. The ants communicate with and influence others using pheromone, a volatile chemical substance. When an ant finds a food source, it deposits certain amount of pheromone along the path and the amount is positively correlated with the quality of the food source. The pheromone laid down biases the path selection of other ants, providing positive feedback. Using the scheme of positive feedback, the algorithm leads the ants to find the shortest path to a best food source \cite{Dorigo1990OptimizationLearningand}.

PSO is motivated by the movement of organisms as a group, as in a flock of birds or a school of fishes. 
The group is represented by a swarm of particles and PSO uses their positions in the search space to represent the feasible solutions of the optimization problem. PSO manipulates the movement of these particles to perform optimization, utilizing the information of individual experience and socio-cognitive tendency. These two kinds of information correspond to cognitive learning and social learning, respectively, and lead the population to find a best way to perform optimization \cite{KennedyEberhart1995Particleswarmoptimization}.

The above two metaheuristics have been applied to solve a vast range of different problems, e.g. \cite{LiaoMolinaStutzle2012ACOAlgorithmBenchmarked}\cite{KirchmaierHaweDiepold2013SwarmIntelligenceinspired}. Motivated by such success, swarm intelligence algorithm design has attracted many researchers and several new algorithms were devised. The most widely studied organism in swarm intelligence is the bee \cite{ParpinelliLopes2011Newinspirationsin}. Abbass proposed a Marriage in honey Bees Optimization (MBO) in \cite{Abbass2001MBO:marriagein} and this algorithm was applied to solve propositional satisfiability problems (3-SAT problems). In MBO, the mating flight of the queen bee is represented as the transitions in a state space (search space), with the queen probabilistically mating with the drone encountered at each state. The probability of mating is determined by the speed and energy of the queen, and the fitness of the drone. Karaboga and Basturk proposed an Artificial Bee Colony optimization (ABC) in \cite{KarabogaBasturk2007Apowerfulandefficient}. ABC classifies the bees in a hive into three types: ``scout bees'' that randomly fly without guidance, ``employed bees'' that search the neighborhood of their positions, and ``onlooker bees'' that use the population fitness to select a guiding solution for exploitation. The algorithm balances exploration and exploitation by means of using employed and onlooker bees for local search, and the scout bees for global search. It also demonstrates satisfactory performance in applications \cite{OmkarSenthilnathKhandelwalNaikGopalakrishnan2011ArtificialBeeColony}\cite{Akay2013studyparticleswarm}.

Besides the bees, other organisms have also been widely studied \cite{ParpinelliLopes2011Newinspirationsin}. Krishnanand and Ghose proposed a Glow-worm Swarm Optimization (GSO) \cite{KrishnanandGhose2005Detectionmultiplesource} based on the behavior of the firefly. In GSO, each firefly randomly selects a neighbor according to its luminescence and moves toward it. In general the fireflies are more likely to get interested in others that glow brighter. As the movement is only conducted locally using selective neighbor information, the firefly swarm is able to divide into disjoint subgroups to explore multiple optima. Another firefly-based technique is proposed by Yang \textit{et al.} \cite{Yang2008Natureinspiredmetaheuristic}. He reformulated the co-movement pattern of fireflies and employed it in optimization. Passino devised a Bacterial Foraging Optimization (BFO) \cite{Passino2002Biomimicrybacterialforaging} based on the bacterial chemotaxis. In BFO, possible solutions to the optimization problem are represented by a colony of bacteria. It consists of three schemes, i.e., chemotaxis, reproduction, and elimination-dispersal. The exploitation task is performed using the first two schemes and the last one contributes to exploration \cite{FarhoodneaMohamedShareefZayandehroodi2014Optimumplacementactive}. Researchers have also devised swarm intelligence algorithms based on other organisms and they can also generate satisfactory optimization performance \cite{ParpinelliLopes2011Newinspirationsin}. To the best of our knowledge, only one spider-inspired metaheuristic aiming at solving optimization problem has been proposed, i.e. the Social Spider Optimization \cite{CuevasCienfuegosZaldivarPerez-Cisneros2013swarmoptimizationalgorithm} devised by Cuevas \textit{et al.}, which divides the spiders into different genders and mimics the mating behavior for optimization. However, our proposed algorithm is totally different from this algorithm in their biological backgrounds, motivations, implementations, and search behaviors. We will further reveal the differences in Section \ref{sub:difference}.

\section{Social Spider Algorithm}\label{sec:ssa}

In SSA, we formulate the search space of the optimization problem as a hyper-dimensional spider web. Each position on the web represents a feasible solution to the optimization problem and all feasible solutions to the problem have corresponding positions on this web. The web also serves as the transmission media of the vibrations generated by the spiders. Each spider on the web holds a position and the quality (or fitness) of the solution is based on the objective function, and represented by the potential of finding a food source at the position. The spiders can move freely on the web. However, they can not leave the web as the positions off the web represent infeasible solutions to the optimization problem. When a spider moves to a new position, it generates a vibration which is propagated over the web. Each vibration holds the information of one spider and other spiders can get the information upon receiving the vibration.

\subsection{Spider}

The spiders are the agents of SSA to perform optimization. At the beginning of the algorithm, a pre-defined number of spiders are put on the web. Each spider $s$ holds a memory, storing the following individual information:
\begin{itemize}
\item The position of $s$ on the web.
\item The fitness of the current position of $s$.
\item The target vibration of $s$ in the previous iteration.
\item The number of iterations since $s$ has last changed its target vibration.
\item The movement that $s$ performed in the previous iteration.
\item The dimension mask\footnote{The dimension mask is a 0-1 binary vector of length D, where D is the dimension of the optimization problem} that $s$ employed to guide movement in the previous iteration.
\end{itemize}
The first two types of information describe the individual situation of $s$, while all others are involved in directing $s$ to new positions. The detailed scheme of movement will be elaborated in Section \ref{sub:search}.

Based on observations, spiders are found to have very accurate senses of vibration. Furthermore, they can separate different vibrations propagated on the same web and sense their respective intensities \cite{Uetz1992Foragingstrategiesspiders}. In SSA, a spider will generate a vibration when it reaches a new position different from the previous one. The intensity of the vibration is correlated with the fitness of the position. The vibration will propagate over the web and other spiders can sense it. In such a way, the spiders on the same web share their personal information with others to form a collective social knowledge.

\subsection{Vibration}\label{sub:vibration}

Vibration is a very important concept in SSA. It is one of the main characteristics that distinguish SSA from other metaheuristics. In SSA, we use two properties to define a vibration, namely, the source position and the source intensity of the vibration. The source position is defined by the search space of the optimization problem, and we define the intensity of a vibration in the range $[0,+\infty)$. Whenever a spider moves to a new position, it generates a vibration at its current position. We define the position of spider $a$ at time $t$ as $\boldsymbol{P}_a(t)$, or simply as $\boldsymbol{P}_a$ if the time argument is $t$. We further use $I(\boldsymbol{P}_a,\boldsymbol{P}_b,t)$ to represent the vibration intensity sensed by a spider at position $\boldsymbol{P}_b$ at time $t$ and the source of the vibration is at position $\boldsymbol{P}_a$. With these notations we can thus use $I(\boldsymbol{P}_s,\boldsymbol{P}_s,t)$ to represent the intensity of the vibration generated by spider $s$ at the source position. This vibration intensity at the source position is correlated with the fitness of its position $f(\boldsymbol{P}_s)$, and we define the intensity value as follows:
\begin{equation}\label{eqn:intensity}
    I(\boldsymbol{P}_s,\boldsymbol{P}_s,t)=\log(\frac{1}{f(\boldsymbol{P}_s) - C}+1)
\end{equation}
where $C$ is a confidently small constant such that all possible fitness values are larger than $C$. Please note that we consider minimization problems in this paper. The design of (\ref{eqn:intensity}) takes the following issues into consideration:

\begin{itemize}
\item All possible vibration intensities of the optimization problem are positive.
\item The positions with better fitness values, i.e. smaller values for minimization problems, have larger vibration intensities than those with worse fitness values.
\item When a solution approaches the global optimum, the vibration intensity would not increase excessively, and cause malfunctioning of the vibration attenuation scheme.
\end{itemize}

As a form of energy, vibration attenuates over distance. This physical phenomenon is accounted for in the design of SSA. We define the distance between spider $a$ and $b$ as $D(\boldsymbol{P}_a,\boldsymbol{P}_b)$ and we use 1-norm (Manhattan distance) to calculate the distance, i.e.,
\begin{equation}\label{eqn:distance}
    D(\boldsymbol{P}_a,\boldsymbol{P}_b)=||\boldsymbol{P}_a-\boldsymbol{P}_b||_1.
\end{equation}
The standard deviation of all spider positions along each dimension is represented by $\boldsymbol{\sigma}$. With these definitions, we further define the vibration attenuation over distance as follows:
\begin{equation}\label{eqn:vibrattendist}
    I(\boldsymbol{P}_a,\boldsymbol{P}_b,t)=I(\boldsymbol{P}_a,\boldsymbol{P}_a,t)\times\exp(-\frac{D(\boldsymbol{P}_a,\boldsymbol{P}_b)}{\overline{\boldsymbol{\sigma}}\times r_a}).
\end{equation}

In the above formula we introduce a user-controlled parameter $r_a\in(0,\infty)$. This parameter controls the attenuation rate of the vibration intensity over distance. The larger $r_a$ is, the weaker the attenuation imposed on the vibration.

\subsection{Search Pattern}\label{sub:search}

Here we demonstrate the above ideas in terms of an algorithm. There are three phases in SSA: initialization, iteration, and final. These three phases are executed sequentially. In each run of SSA, we start with the initialization phase, then perform searching in an iterative manner, and finally terminate the algorithm and output the solutions found.

In the initialization phase, the algorithm defines the objective function and its solution space. The value for the parameter used in SSA is also assigned. After setting the values, the algorithm proceeds to create an initial population of spiders for optimization. As the total number of spiders remains unchanged during the simulation of SSA, a fixed size memory is allocated to store their information. The positions of spiders are randomly generated in the search space, with their fitness values calculated and stored. The initial target vibration of each spider in the population is set at its current position, and the vibration intensity is zero. All other attributes stored by each spider are also initialized with zeros. This finishes the initialization phase and the algorithm starts the iteration phase, which performs the search with the artificial spiders created.

In the iteration phase, a number of iterations are performed by the algorithm. In each iteration, all spiders on the web move to a new position and evaluate their fitness values. Each iteration can be further divided into the following sub-steps: fitness evaluation, vibration generation, mask changing, random walk, and constraint handling.

The algorithm first calculates the fitness values of all the artificial spiders on different positions on the web, and update the global optimum value if possible. The fitness values are evaluated once for each spider during each iteration. Then these spiders generate vibrations at their positions using (\ref{eqn:intensity}). After all the vibrations are generated, the algorithm simulates the propagation process of these vibrations using (\ref{eqn:vibrattendist}). In this process, each spider $s$ will receive $|pop|$ different vibrations generated by other spiders where $pop$ is the spider population. The received information of these vibrations include the source position of the vibration and its attenuated intensity. We use $V$ to represent these $|pop|$ vibrations. Upon the receipt of $V$, $s$ will select the strongest vibration $v_\textit{s}^\textit{best}$ from $V$ and compare its intensity with the intensity of the target vibration $v_\textit{s}^\textit{tar}$ stored in its memory. $s$ will store $v_\textit{s}^\textit{best}$ as $v_\textit{s}^\textit{tar}$ if the intensity of $v_\textit{s}^\textit{best}$ is larger, and $c_s$, or the number of iterations since $s$ has last changed its target vibration, is reset to zero; otherwise, the original $v_\textit{tar}$ is retained and $c_s$ is incremented by one. We use $\boldsymbol{P}_\textit{s}^\textit{i}$ and $\boldsymbol{P}_\textit{s}^\textit{tar}$ to represent the source positions of $V$ and $v_\textit{tar}$, respectively, and $i=\{1,2,\cdots,|pop|\}$.

The algorithm then manipulates $s$ to perform a random walk towards $v_\textit{s}^\textit{tar}$. Here we utilize a dimension mask to guide the movement. Each spider holds a dimension mask $\boldsymbol{m}$, which is a 0-1 binary vector of length $D$ and $D$ is the dimension of the optimization problem. Initially all values in the mask are zero. In each iteration, spider $s$ has a probability of $1 - {p_c}^{c_s}$ to change its mask where $p_c\in(0,1)$ is a user-defined attribute that describes the probability of changing mask. If the mask is decided to be changed, each bit of the vector has a probability of $p_m$ to be assigned with a one, and $1-p_m$ to be a zero. $p_m$ is also a user-controlled parameter defined in $(0,1)$. Each bit of a mask is changed independently and does not have any correlation with the previous mask. In case all bits are zeros, one random value of the mask is changed to one. Similarly, one random bit is assigned to zero if all values are ones.

After the dimension mask is determined, a new following position $\boldsymbol{P}_\textit{s}^\textit{fo}$ is generated based on the mask for $s$. The value of $i$-th dimension of the following position $\boldsymbol{P}_\textit{s,i}^\textit{fo}$ is generated as follows.
\begin{equation}\label{eqn:follow}
    P_\textit{s,i}^\textit{fo} = 
    \begin{cases}
    P_\textit{s,i}^\textit{tar} & m_{s,i} = 0 \\
    P_\textit{s,i}^\textit{r} & m_{s,i} = 1 \\
    \end{cases},
\end{equation}
where $r$ is a random integer value generated in $[1, |pop|]$, and $m_{s,i}$ stands for the $i$-th dimension of the dimension mask $\boldsymbol{m}$ of spider $s$. Here the random number $r$ for two different dimensions with $m_{s,i} = 1$ is generated independently.

With the generated $\boldsymbol{P}_\textit{s}^\textit{fo}$, $s$ performs a random walk to the position. This random walk is conducted using the following equation.
\begin{equation}\label{eqn:walk}
    \boldsymbol{P}_s(t+1)=\boldsymbol{P}_s+(\boldsymbol{P}_s - \boldsymbol{P}_s(t-1))\times r + (\boldsymbol{P}_\textit{s}^\textit{fo} - \boldsymbol{P}_s)\odot\boldsymbol{R},
\end{equation}
where $\odot$ denotes element-wise multiplication and $\boldsymbol{R}$ is a vector of random float-point numbers generated from zero to one uniformly. Before following $\boldsymbol{P}_\textit{s}^\textit{fo}$, $s$ first moves along its previous direction, which is the direction of movement in the previous iteration. The distance along this direction is a random portion of the previous movement. Then $s$ approaches $\boldsymbol{P}_\textit{s}^\textit{fo}$ along each dimension with random factors generated in $(0,1)$. This random factor for different dimensions are generated independently. After this random walk, $s$ stores its movement in the current iteration for the next iteration. This ends the random walk sub-step.

The final sub-step of the iteration phase is the constraint handling. The spiders may move out of the web during the random walk step, which causes the constraints of the optimization problem to be violated.  There are many methods to handle the boundary constraints in the previous literature, and the random approach, absorbing approach, and the reflecting approach are three most widely-adopted methods \cite{ChuGaoSorooshian2011Handlingboundaryconstraints}. In this paper we adopt the reflecting approach for constraint handling and produce a boundary-constraint-free position $\boldsymbol{P}_s(t+1)$ by
\begin{equation}\label{eqn:constraint}
    P_{s,i}(t+1) =
    \begin{cases}
    (\overline{x_i} - P_{s,i}) \times r & \mathrm{if\:}P_{s,i}(t+1) > \overline{x_i} \\
    (P_{s,i} - \underline{x_i}) \times r & \mathrm{if\:}P_{s,i}(t+1) < \underline{x_i} \\
    \end{cases},
\end{equation}
where $\overline{x_i}$ is the upper bound of the search space in the $i$-th dimension, and $\underline{x_i}$ is the lower bound of the corresponding dimension. $r$ is a random floating point number generated in $(0, 1)$

The iteration phase loops until the stopping criteria is matched. The stopping criteria can be defined as the maximum iteration number reached, the maximum CPU time used, the error rate reached, the maximum number of iterations with no improvement on the best fitness value, or any other appropriate criteria. After the iteration phase, the algorithm outputs the best solution with the best fitness found. The above three phases constitute the complete algorithm of SSA and its pseudo-code can be found in Algorithm \ref{alg:ssa}.

\begin{algorithm}
\caption{\sc{Social Spider Algorithm}}
  \begin{algorithmic}[1]
  \State Assign values to the parameters of SSA.
  \State Create the population of spiders $pop$ and assign memory for them.
  \State Initialize $v_\textit{s}^\textit{tar}$ for each spider.
  \While {stopping criteria not met}
    \For {\textbf{each} spider $s$ in $pop$}
      \State Evaluate the fitness value of $s$.
      \State Generate a vibration at the position of $s$.
    \EndFor
    \For {\textbf{each} spider $s$ in $pop$}
      \State Calculate the intensity of the vibrations $V$ \par\hspace{2.1em} generated by all spiders.
      \State Select the strongest vibration $v_\textit{s}^\textit{best}$ from $V$.
      \If {The intensity of $v_\textit{s}^\textit{best}$ is larger than $v_\textit{s}^\textit{tar}$}
        \State Store $v_\textit{s}^\textit{best}$ as $v_\textit{s}^\textit{tar}$.
      \EndIf
      \State Update $c_s$.
      \State Generate a random number $r$ from [0,1).
      \If {$r > {p_c}^{c_s}$}
        \State Update the dimension mask $\boldsymbol{m}_s$.
      \EndIf
      \State Generate $\boldsymbol{P}_\textit{s}^\textit{fo}$.
      \State Perform a random walk.
      \State Address any violated constraints.
    \EndFor
  \EndWhile
  \State Output the best solution found.
  \end{algorithmic}
  \label{alg:ssa}
\end{algorithm}

\subsection{Differences between SSA and Other Evolutionary Computation Algorithms}\label{sub:difference}

A number of swarm intelligence algorithms have been proposed in the past few decades. Among them, PSO and ACO are the two most widely employed and studied algorithms. SSA may also be classified as a swarm intelligence algorithm, but it has many differences from PSO and ACO, elaborated below.

PSO, like SSA, was originally proposed for solving continuous optimization problems. It was also inspired by animal behavior. However, the first crucial difference between SSA and PSO is in individual following patterns. In PSO, all particles follow a common global best position and their own personal best position. However in SSA, all spiders follow positions constructed by others' current positions and their own historical positions. These following positions are not guaranteed to be visited by the population before, and different spiders can have different following positions. Since the global best position and spiders' current positions differ greatly during most time of the optimization process, these two following patterns lead to different searching behaviors. This may weaken the convergence ability of SSA but can potentially strengthen the capability of solving multi-modal optimization problems with a great number of local optimums.

Besides the difference in the following pattern, the difference in their biology backgrounds is also very significant. PSO was designed based on the model of coordinated group animal motions of flocks of birds or schools of fishes. This model serves as the design metaphor of PSO. SSA is inspired by the social spider foraging strategy, which belongs to the scope of general social animal searching behavior. We use a general IS model as the design framework. This difference is also a major distinguishing feature of SSA from other proposed algorithms.

A third difference between SSA and the original formulation of PSO is in the information propagation method. In PSO, the information propagation method is neglected, and each particle is assumed to be aware of all the information of the system without loss. Although the information validity ranges are considered in some recent variants of PSO, the information loss characteristic is still a unique feature that distinguishes SSA from PSO variants. In SSA we model the information propagation process through the vibrations on the spider web. This process forms a general knowledge system with information loss. Although there is still no research on how the information loss will impact the social foraging strategy employed in optimization, it is possible that this information loss system partially contributes to the performance improvement of SSA over PSO.

Another difference is that in PSO, the common knowledge of the group is all about the best particle in the system. All remaining particles in the system do not constitute the shared information of the group, which may lead to neglecting some valuable information of the population. In SSA, each spider generates a new piece of information and propagates the information to the whole population. There are also some PSO variants that shares personal best position information with the population, but the main focus of PSO is on the best positions of the individuals and the population. In SSA, the information generated and propagated with the vibrations are the current positions instead of the best-in-history positions, which may differ greatly with the search.

Although both SSA and ACO draw their inspirations from the social animal foraging strategy, there are still some obvious differences. The foraging frameworks adopted by the two algorithms are quite different: ACO utilizes the ant foraging behavior to perform optimization. Ants find food by laying down pheromone trails and collectively establishing positive feedbacks which bias the later path selection, while spiders sense the vibration propagated by the spider web to locate the prey. Another difference is the presentation of feasible solutions. In SSA we use the positions on the spider web to represent feasible solutions. Similar representations have also been widely adopted in the swarm intelligence algorithm. Meanwhile, ACO uses the path between the ant hive and food sources to represent solutions to the optimization problems. Additionally, ACO was originally designed to solve combinatorial problems. Although in recent years there are ACO-variant algorithms designed mainly to solve continuous problems \cite{SochaDorigo2008Antcolonyoptimization}, the performance is not as good as the performance of the original ACO in solving combinatorial problems like the Traveling Salesman Problem. There are also information propagation and searching pattern differences between SSA and ACO as described above.

There are also some other swarm intelligence algorithms proposed to solve continuous problems, and SSA has some unique characteristics. In most swarm intelligence algorithms, e.g., ABC and GSO, the populations are structured into different types. Different types of individuals perform different jobs and the whole population cooperates to search the solution space. However in SSA, all individuals (spiders) are equal. Each performs all the tasks that would be executed by multiple types of the populations in other algorithms. If we put SSA into the conventional framework, it has the feature that the different types of individuals can transform into other types very smoothly and without the guidance of the user, which may potentially contribute to the performance improvement.

As to Social Spider Optimization (SSO) \cite{CuevasCienfuegosZaldivarPerez-Cisneros2013swarmoptimizationalgorithm}, differences lie in all aspects of the algorithm design. A most important difference is that in SSO the spiders are classified by gender. Male and female spiders have different searching operations. However, the spiders in SSA share the same searching operation, significantly reducing the effort in implementation. SSA also incorporates the information propagation model into its algorithm design, and thus the social spider population in SSA fits the IS model. Besides, SSA imitates the foraging behavior of social spiders, while SSO imitates the mating behavior of social spiders. The differences in algorithm implementation are more patent. In SSO there are three spider movement operators executed first in parallel and then in sequence. The moving pattern of the third operator highly depends on the first two operators. This design may potentially increase the difficulty of analyzing the search behavior of the algorithm. In SSA we implement one random move operator, which combines both exploration and exploitation behaviors in one move. In our design, the search behavior is controlled by the parameters, thus providing a clear view on the search behavior of the algorithm. The impact of different parameters on the optimization performance of SSA is further illustrated in Section \ref{sub:param}.

Although EAs, like GA and ES, are also population-based algorithms, and inevitably share some similarities with the population-based SSA, they are quite different general-purpose metaheuristics. They are inspired by completely different biological disciplines. EAs usually employ different recombination and decomposition operators to manipulate the solutions, which imitate the regeneration of an organism.

As stated above, although we still do not know the exact impact of information loss on the optimization process, this feature of SSA may contribute to the optimum search in some complex multimodal optimization problems. The uniform structure of the population is another potential advantage of SSA. In addition, the unique searching pattern and its underlying social animal foraging strategy as well as the IS foraging model contribute to the overall performance of SSA.

\subsection{Adjusting SSA Parameters}\label{sub:param}

Choosing proper parameters of SSA for numerical and real-world optimization problems can be time-consuming. The trial-and-error scheme, or a parameter sweep test, may reveal the best achievable performance over the parameter space at the expense of high computational cost. In real-world optimization problems, evaluating the fitness function may take a long time, much longer than evaluating our benchmark functions, and one evaluation may take several seconds or even minutes to finish, rendering trial-and-error schemes impractical for parameter tuning. As alternatives, researchers have proposed some schemes to replace the trial-and-error parameter selection scheme. These schemes can generally be classified into three groups \cite{QinLi2013DifferentialEvolutionCEC}:
\begin{itemize}
\item \textit{Fixed parameter schemes} select a parameter combination before the simulation using empirical or theoretical knowledge of the characteristics of the parameters. This combination remains constant throughout the whole search \cite{LamLiYu2012RealCodedChemical}\cite{PriceStornLampinen2005DifferentialEvolutionPractical}.
\item \textit{Deterministic parameter schemes} use some pre-defined rules to change the parameter values throughout the search \cite{LamLiYu2012RealCodedChemical}\cite{ChenZhangLinChenZhanChungLiShi2013ParticleSwarmOptimization}.
\item \textit{Adaptive parameter schemes} change the parameter values by adaptively learning the impact of changing parameters on the searching performance throughout the search \cite{QinHuangSuganthan2009DifferentialEvolutionAlgorithm}. Some schemes encode the parameters into the solution and evolve the parameters together with the population \cite{VrugtRobinsonHyman2009SelfAdaptiveMultimethod}.
\end{itemize}

In this paper we use the fixed parameter scheme to test the performance of SSA compared with other algorithms. We also use this scheme to perform a preliminary parameter sensitivity analysis in order to deduce some rules of thumb on choosing parameters that can consistently lead to satisfactory results on a wide range of functions with different characteristics. This test can also discover some of the features of the parameters when solving different kinds of optimization problems. We carry out extensive simulations on our benchmark functions, which cover a wide range of optimization problems. Thus, the derived rules of thumb can be expected to give generally good performance on unknown problems.

In SSA we employ three user-controlled parameters to guide the searching behaviour, namely,

\begin{itemize}
\item $r_a$: This parameter defines the rate of vibration attenuation when propagating over the spider web.
\item $p_c$: This parameter controls the probability of the spiders changing their dimension mask in the random walk step.
\item $p_m$: This parameter defines the probability of each value in a dimension mask to be one.
\end{itemize}

\begin{figure*}
  \centering \includegraphics[width=\linewidth]{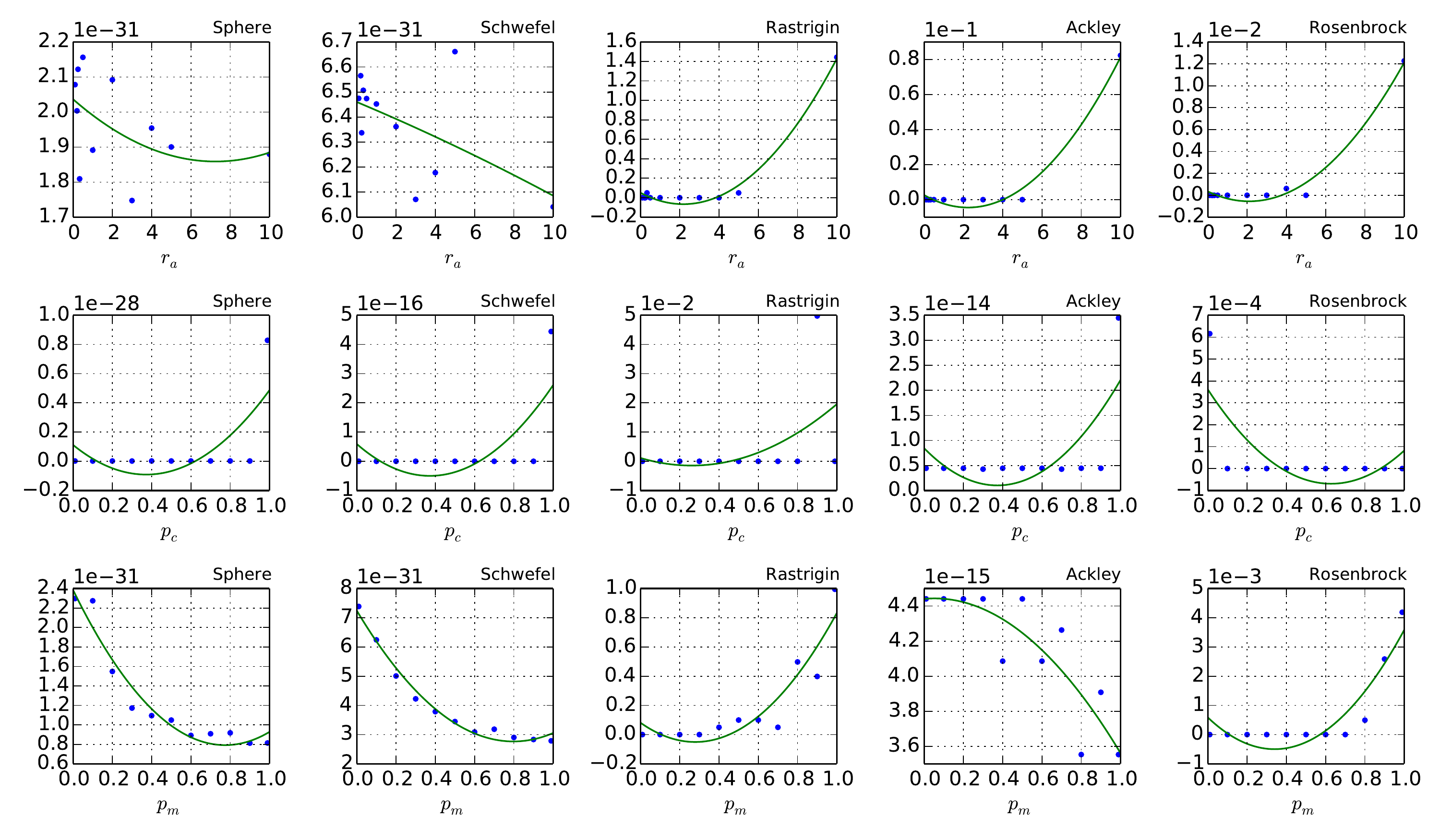}
  \caption{Parameter test results on $r_a$, $p_c$,  and $p_m$.}
  \label{fig:parameter}
\end{figure*}

In this section, we employ five 10-dimensional benchmark functions to investigate the impact of these parameters on the performance of SSA. These functions are the Sphere, Schwefel 2.22, Rastrigin, Ackley, and Griewank functions and the detailed definition of these functions can be found in Section \ref{sec:benchmark}. All benchmark functions are not shifted. The value of $r_a$ is selected from the set $\{\frac{1}{10},\frac{1}{5},\frac{1}{4},\frac{1}{3},\frac{1}{2},1,2,3,4,5,10\}$, and the values of $p_c$ and $p_m$ are both selected from the set $\{0.01, 0.1, 0.2, 0.3, 0.4, 0.5, 0.6, 0.7, 0.8, 0.9, 0.99\}$. So for each function/parameter pair there are 11 data points for analysis. The stopping criteria is set to 100 000 evaluations and each function is tested for 20 times. The population size is set to 10 in accordance with our later simulation/ The mean results are plotted in Fig. \ref{fig:parameter} with dots, and the second-order polynomial regression curve for each function is also plotted for demonstration. As the Sphere and Schwefel 2.22 functions are uni-modal functions while the other three are multi-modal functions, we can obtain some interesting observations from the mean results and the regression curve.

From Fig. \ref{fig:parameter} we can see SSA is very robust in solving unimodal problems when the three parameters change. All six sub-figures indicate outstanding performance in terms of fitness value. A general conclusion is that the value of $p_c$ shall not be set to a very large value, e.g. 0.99.

When considering the multimodal problems, we can observe some obvious impact of the parameters on the performance. In terms of the vibration attenuation rate $r_a$, both the mean results and the regression curves favor relatively small values. A generally preferred value is one, while random values selected from $(0,3]$ shall also be able to yield good results.

While $r_a$ presents relatively stable results in the multimodal problem tests, $p_c$ has different impacts on different problems. Rastrigin function favors a small $p_c$ while Rosenbrock function prefers a medium-large one. It seems that most $p_c$ values can generate good results on Ackley function. So the selection of $p_c$ highly depends on the nature of the optimization problems to be solved.

As to $p_m$, the simulation results reveal a similar tendency with $r_a$ that the multimodal problems seem to prefer a relatively small $p_m$ somewhere near 0.1. So we adopt the parameter combination $r_a = 1$, $p_c = 0.7$, and $p_m = 0.1$ in all of our later simulations. Please note that this parameter combination is not guaranteed to be the best one for solving all optimization problems, and parameter tuning is essential to address unfamiliar problems.

Please note that the parameter sensitivity analysis in this paper is a preliminary one. In this test only one parameter is tested while the remaining two is set unchanged, i.e., $r_a = 1$, $p_c = 0.7$, and $p_m = 0.1$. Although it is not guaranteed that this set of parameters works best for the benchmark functions, it is one that yields outstanding performance, which will be demonstrated in Section \ref{sec:benchmark}. A complete parameter sensitivity analysis is one of the future research topics of SSA.

\section{Benchmark Problems and Evaluation Method}\label{sec:benchmark}

In order to benchmark the performance of SSA, we conduct simulations on 25 different benchmark functions. These benchmark functions are all the base functions from the latest Competition on Real-Parameter Single Objective Optimization Problems at CEC 2013 \cite{LiangQuSuganthanHernandez-Diaz2013ProblemDefinitionsand} and CEC 2014 \cite{LiangQuSuganthan2014ProblemDefinitionsand}. The benchmark functions can be classified into four groups:
\begin{itemize}
\item Group I: $f_1$--$f_5$ are unimodal functions.
\item Group II: $f_6$--$f_{15}$ are multimodal functions.
\item Group III: $f_{16}$--$f_{20}$ are rotated multimodal functions whose base functions belong to Group II functions.
\item Group IV: $f_{21}$--$f_{25}$ are hybrid multimodal functions whose base functions belong to Group I -- III functions.
\end{itemize}
The benchmark functions are listed in Table \ref{tbl:benchmark}. All benchmark functions, except $f_{13}$ Schwefel's Problem 2.26, are shifted minimization problems and the search ranges are scaled to $[-100,100]^n$, where $n$ is the dimension of the problem. Group I functions are used to test the fast-converging performance of SSA. Group II functions all have a large number of local minima points, and can be used to test the ability of SSA to jump out of local optima and avoid pre-mature convergence. Group III functions are more complex than other functions and can push the searching capability of SSA to a limit. Group IV functions are employed to test the optimization performance of handling problems consisting of different subcomponents with different properties. The detailed implementation of Group IV functions, i.e. hybrid multimodal functions, can be found in \cite{LiangQuSuganthan2014ProblemDefinitionsand}. In Table \ref{tbl:benchmark} we only list the subcomponents of the hybrid functions.

\begin{table*}[t]
  \caption{Benchmark Functions}
  \label{tbl:benchmark}
  \tiny
  \begin{center}
    \begin{threeparttable}
      \begin{tabular}{lll}
        \hline
        Function & Transformation\tnote{*} & Name \\
		\hline
        $\begin{aligned}f_1(\boldsymbol{z}) = \sum\nolimits_{i=1}^nz_i^2\end{aligned}$ & $\boldsymbol{z} = \boldsymbol{x} - \boldsymbol{o}$ & Sphere Function \\ 
        $\begin{aligned}f_2(\boldsymbol{z}) = \sum\nolimits_{i=1}^n|z_i|+\prod\nolimits_{i=1}^n|z_i|\end{aligned}$ & $\boldsymbol{z} = (\boldsymbol{x} - \boldsymbol{o})\times10/100$ & Schwefel's Problem 2.22 \\ 
        $\begin{aligned}f_3(\boldsymbol{z}) = z_1^2+10^6\sum\nolimits_{i=2}^nz_i^2\end{aligned}$ & $\boldsymbol{z} = \boldsymbol{x} - \boldsymbol{o}$ & Cigar Function \\
        $\begin{aligned}f_4(\boldsymbol{z}) = 10^6z_1^2+\sum\nolimits_{i=2}^nz_i^2\end{aligned}$ & $\boldsymbol{z} = \boldsymbol{x} - \boldsymbol{o}$ & Discus Function \\
        $\begin{aligned}f_5(\boldsymbol{z}) = \sum\nolimits_{i=1}^nix_i^4+rand()\tnote{+}\end{aligned}$ & $\boldsymbol{z} = (\boldsymbol{x} - \boldsymbol{o})\times1.28/100$ & Quadratic Function with Noise \\
		$\begin{aligned}f_6(\boldsymbol{z})=\sum\nolimits_{i=1}^n(z_i^2-10\cos(2\pi z_i)+10)\end{aligned}$ & $\boldsymbol{z} = (\boldsymbol{x} - \boldsymbol{o})\times5.12/100$ & Rastrigin Function \\
		$\begin{aligned}f_7(\boldsymbol{z})=-20\exp(-0.2\sqrt{\frac{1}{n}\sum\nolimits_{i=1}^nz_i^2})-\exp[\frac{1}{n}\sum\nolimits_{i=1}^n\cos(2\pi z_i)]+20+e\end{aligned}$ & $\boldsymbol{z} = (\boldsymbol{x} - \boldsymbol{o})\times32/100$ & Ackley Function \\
		$\begin{aligned}f_8(\boldsymbol{z}) = \frac{1}{4000}\sum\nolimits_{i=1}^nz_i^2-\prod_{i=1}^n\cos(\frac{z_i}{\sqrt{i}})+1\end{aligned}$ & $\boldsymbol{z} = (\boldsymbol{x} - \boldsymbol{o})\times600/100$ & Griewank Function \\
		$\begin{aligned}f_9(\boldsymbol{z})=\sum\nolimits_{i=1}^{n-1}(100(z_{i+1}-z_{i}^2)^2+(z_i-1)^2)\end{aligned}$ & $\boldsymbol{z} = (\boldsymbol{x} - \boldsymbol{o})\times30/100$ & Rosenbrock Function \\
		$\begin{aligned}f_{10}(\boldsymbol{z})=&\sin^2(\pi y_1)+\sum\nolimits_{i=1}^{n-1}[(y_i-1)^2(1+10(\sin^2y_{i+1}))]+\\&(y_n-1)^2(1+\sin^2(2\pi y_n)), y_i = 1+\frac{1}{4}(z_i+1) \end{aligned}$ & $\boldsymbol{z} = (\boldsymbol{x} - \boldsymbol{o})\times50/100$ & Levy Function \\
		$\begin{aligned}f_{11}(\boldsymbol{z})=&\frac{1}{10}[\sin^2(3\pi z_1)+\sum\nolimits_{i=1}^{n-1}(z_i-1)^2(1+\sin^2(3\pi z_{i+1}))+\\&(z_n-1)^2(1+\sin^2(2\pi z_n))]+\sum\nolimits_{i=1}^n u(z_i,5,100,4) \\&u(z_i,a,k,m)=\begin{cases}k(z_i-a)^m & \mathrm{for\:}z_i>a\\0& \mathrm{for\:}-a\leq z_i\leq a\\k(-z_i-a)^m & \mathrm{for\:}z_i<-a \end{cases}\end{aligned}$ & $\boldsymbol{z} = (\boldsymbol{x} - \boldsymbol{o})\times50/100$ & Penalized Function \\
		$\begin{aligned}f_{12}(\boldsymbol{z})=&g(z_1,z_2)+g(z_2,z_3)+\cdots+g(z_{n-1},z_n)+g(z_n,z_1)\\&g(x,y)=0.5+\frac{(\sin^2(\sqrt{x^2+y^2})-0.5)}{(1+0.001(x^2+y^2))^2}\end{aligned}$ & $\boldsymbol{z} = \boldsymbol{x} - \boldsymbol{o}$ & Schaffer's Function F6 \\
		$\begin{aligned}f_{13}(\boldsymbol{z})=418.9828872724338 * n-\sum\nolimits_{i=1}^n(z_i\sin\sqrt{|z_i|}) \end{aligned}$ & $\boldsymbol{z} = \boldsymbol{x}\times500/100$ & Schwefel's Problem 2.26 \\
		$\begin{aligned}f_{14}(\boldsymbol{z})=[\frac{1}{n-1}\sum\nolimits_{i=1}^{n-1}(\sqrt{y_i}+\sin(50y_i^{0.2})\sqrt{y_i})]^2,y_i=\sqrt{z_i^2+z_{i+1}^2}\end{aligned}$ & $\boldsymbol{z} = \boldsymbol{x} - \boldsymbol{o}$ & Schaffer's Function F7 \\
		$\begin{aligned}f_{15}(\boldsymbol{z})=&\min(\sum\nolimits_{i=1}^n(z_i-\mu_1)^2, d\times n+s\times\sum\nolimits_{i=1}^n(z_i-\mu_2)^2)+\\&10\sum\nolimits_{i=1}^n(1-\cos[2\pi(z_i-\mu_1)])\\&s=1-\frac{1}{2\sqrt{n}-8.2},\mu_1=2.5,\mu_2=-\sqrt{\frac{\mu_1^2-1}{s}}\end{aligned}$ & $\boldsymbol{z} = (\boldsymbol{x} - \boldsymbol{o})\times10/100$ & Lunacek Function \\
		$\begin{aligned}f_{16}(\boldsymbol{z})=f_8(\boldsymbol{Mz})\end{aligned}$ & $\boldsymbol{z} = (\boldsymbol{x} - \boldsymbol{o})\times600/100$ & Rotated Griewank Function \\
		$\begin{aligned}f_{17}(\boldsymbol{z})=f_9(\boldsymbol{Mz})\end{aligned}$ & $\boldsymbol{z} = (\boldsymbol{x} - \boldsymbol{o})\times30/100$ & Rotated Rosenbrock Function \\
		$\begin{aligned}f_{18}(\boldsymbol{z})=f_{11}(\boldsymbol{Mz})\end{aligned}$ & $\boldsymbol{z} = (\boldsymbol{x} - \boldsymbol{o})\times50/100$ & Rotated Penalized Function \\
		$\begin{aligned}f_{19}(\boldsymbol{z})=f_{12}(\boldsymbol{Mz})\end{aligned}$ & $\boldsymbol{z} = \boldsymbol{x} - \boldsymbol{o}$ & Rotated Schaffer's Function F6 \\
		$\begin{aligned}f_{20}(\boldsymbol{z})=f_{15}(\boldsymbol{Mz})\end{aligned}$ & $\boldsymbol{z} = (\boldsymbol{x} - \boldsymbol{o})\times10/100$ & Rotated Lunacek Function \\
		$\begin{aligned}f_{21}(\boldsymbol{z})=f_1(\boldsymbol{z_1})+f_6(\boldsymbol{z_2})+f_{13}(\boldsymbol{z_3})\end{aligned}$ & $\boldsymbol{z} = [\boldsymbol{z_1}, \boldsymbol{z_2}, \boldsymbol{z_3}]$ & Hybrid Function 1\\
		$\begin{aligned}f_{22}(\boldsymbol{z})=f_6(\boldsymbol{z_1})+f_8(\boldsymbol{z_2})+f_9(\boldsymbol{z_3})\end{aligned}$ & $\boldsymbol{z} = [\boldsymbol{z_1}, \boldsymbol{z_2}, \boldsymbol{z_3}]$ & Hybrid Function 2\\
		$\begin{aligned}f_{23}(\boldsymbol{z})=f_3(\boldsymbol{z_1})+f_7(\boldsymbol{z_2})+f_9(\boldsymbol{z_3})+f_{11}(\boldsymbol{z_4})\end{aligned}$ & $\boldsymbol{z} = [\boldsymbol{z_1}, \boldsymbol{z_2}, \boldsymbol{z_3}, \boldsymbol{z_4}]$ & Hybrid Function 3\\
		$\begin{aligned}f_{24}(\boldsymbol{z})=f_6(\boldsymbol{z_1})+f_7(\boldsymbol{z_2})+f_8(\boldsymbol{z_3})+f_9(\boldsymbol{z_4})+f_{13}(\boldsymbol{z_5})\end{aligned}$ & $\boldsymbol{z} = [\boldsymbol{z_1}, \boldsymbol{z_2}, \boldsymbol{z_3}, \boldsymbol{z_4}, \boldsymbol{z_5}]$ & Hybrid Function 4\\
		$\begin{aligned}f_{25}(\boldsymbol{z})=f_1(\boldsymbol{z_1})+f_7(\boldsymbol{z_2})+f_{10}(\boldsymbol{z_3})+f_{13}(\boldsymbol{z_4})+f_{15}(\boldsymbol{z_5})\end{aligned}$ & $\boldsymbol{z} = [\boldsymbol{z_1}, \boldsymbol{z_2}, \boldsymbol{z_3}, \boldsymbol{z_4}, \boldsymbol{z_5}]$ & Hybrid Function 5\\
		\hline
        \end{tabular}
        \begin{tablenotes}\footnotesize
            \item [*] $\boldsymbol{o}$ is a shifting vector and $\boldsymbol{M}$ is a transformation matrix. $\boldsymbol{o}$ and $\boldsymbol{M}$ can be obtained from  \cite{LiangQuSuganthan2014ProblemDefinitionsand}.
            \item [+] $rand()$ is a random number uniformly generated in $(0, 1)$.
        \end{tablenotes}
    \end{threeparttable}
  \end{center}
\end{table*}

All benchmark functions locate their global minimum values at zero, and the fitness values smaller than $10^{-8}$ are considered as $10^{-8}$ as required in \cite{LiangQuSuganthanHernandez-Diaz2013ProblemDefinitionsand} and \cite{LiangQuSuganthan2014ProblemDefinitionsand}. We test the benchmark functions in 10, 30, and 50 dimensions to draw empirical conclusion on the scalability of SSA, and each function is tested for 51 runs \cite{LiangQuSuganthan2014ProblemDefinitionsand}. In each run, we use a maximum number of $10^4\times n$ function evaluations as the termination criteria. In order to meet the requirement set by \cite{LiangQuSuganthan2014ProblemDefinitionsand}, we use one fixed combination of parameters for SSA in the simulation of all groups of functions. The population size is $n$, and other parameters are set according to the analysis in Section \ref{sub:param}.

To evaluate the performance of SSA, we compare the simulation results with the state-of-the-art algorithms in solving real-parameter optimization problems, including the variances of Co-variance Matrix Adaptation Evolution Strategies (CMA-ES) \cite{HansenOstermeier2001CompletelyDerandomizedSelf}, adaptive Differential Evolution algorithms (DE) \cite{StornPrice1997Differentialevolution:simple}, and a Global and Local real-coded Genetic Algorithm (GL-25) \cite{Garcia-MartinezLozanoHerreraMolinaSanchez2008Globalandlocal}. CMA-ES and DE variants are arguably the most successful optimization algorithms current in use \cite{DasSuganthan2011DifferentialEvolution:Survey}. In the latest Competition on Real-Parameter Single Objective Optimization Problems at CEC 2013, their variant algorithms possess nine positions in the top ten best performing algorithms. For DE variants, we select JADE \cite{ZhangSanderson2009JADE:adaptivedifferential} and SaDE \cite{QinHuangSuganthan2009DifferentialEvolutionAlgorithm} for comparison due to their excellent performance demonstrated in the CEC 2013 competition. All source codes are obtained from the original author. We make some minor changes to adapt them to our benchmark functions, but the main body and logic of the algorithms are untouched. The stopping criteria for all the compared algorithms are set to $10^4\times n$ function evaluations, and the parameters of these algorithms are set according to the recommendation made in the corresponding literature, i.e., CMA-ES in \cite{Loshchilov2013CMAESwith}, JADE in \cite{ZhangSanderson2009JADE:adaptivedifferential}, SaDE in \cite{QinHuangSuganthan2009DifferentialEvolutionAlgorithm}, and GL-25 in \cite{Garcia-MartinezLozanoHerreraMolinaSanchez2008Globalandlocal}.

Besides these state-of-the-art algorithms, we also performed simulations with other famous algorithms, namely, Real-Coded Genetic Algorithm \cite{Holland1975AdaptationinNatural}, Adaptive Particle Swarm Optimization \cite{ZhanZhangLiChung2009AdaptiveParticleSwarm},  Artificial Bee Colony Optimization \cite{KarabogaBasturk2007Apowerfulandefficient}, Firefly Algorithm \cite{Yang2008Natureinspiredmetaheuristic}, Cuckoo Search \cite{YangDeb2009Engineeringoptimisationby}, and Group Search Optimizer \cite{HeWuSaunders2009GroupSearchOptimizer:}. As their overall performance in terms of best fitness values achieved and convergence speed is not comparable to SSA and the other four algorithms, the detailed simulation results will not be presented in this paper.

\section{Numerical Experiments and Results}\label{sec:result}

In this section we present the simulation results of SSA on the benchmark functions identified in Section \ref{sec:benchmark}. We perform comparison among SSA and other algorithms and give statistical analysis on the simulation results.

\subsection{Experimental Comparison with Other State-of-the-Art Algorithms}\label{sub:compare}

\begin{sidewaystable}[t]
  \caption{Simulation Results for 30-D Problems}
  \label{tbl:comp30d}
  \scriptsize
  \begin{center}
    \begin{threeparttable}
      \begin{tabular}{cccccc}
        \hline
        Function & CMA-ES & JADE & SaDE & GL25 & SSA \\
        & Mean$\pm$Std Dev & Mean$\pm$Std Dev & Mean$\pm$Std Dev & Mean$\pm$Std Dev & Mean$\pm$Std Dev \\
					\hline
$f_1$ & \cellcolor{blue!20}\texttt{1.00E-08$\pm$0.00E+00}$\odot$ & \cellcolor{blue!20}\texttt{1.00E-08$\pm$0.00E+00}$\odot$ & \cellcolor{blue!20}\texttt{1.00E-08$\pm$0.00E+00}$\odot$ & \cellcolor{blue!20}\texttt{1.00E-08$\pm$0.00E+00}$\odot$ & \cellcolor{blue!20}\texttt{1.00E-08$\pm$0.00E+00}\\
$f_2$ & \cellcolor{blue!20}\texttt{1.00E-08$\pm$0.00E+00}$\odot$ & \cellcolor{blue!20}\texttt{1.00E-08$\pm$0.00E+00}$\odot$ & \cellcolor{blue!20}\texttt{1.00E-08$\pm$0.00E+00}$\odot$ & \cellcolor{blue!20}\texttt{1.00E-08$\pm$0.00E+00}$\odot$ & \cellcolor{blue!20}\texttt{1.00E-08$\pm$0.00E+00}\\
$f_3$ & \cellcolor{blue!20}\texttt{1.00E-08$\pm$0.00E+00}$\odot$ & \cellcolor{blue!20}\texttt{1.00E-08$\pm$0.00E+00}$\odot$ & \cellcolor{blue!20}\texttt{1.00E-08$\pm$0.00E+00}$\odot$ & \cellcolor{blue!20}\texttt{1.00E-08$\pm$0.00E+00}$\odot$ & \cellcolor{blue!20}\texttt{1.00E-08$\pm$0.00E+00}\\
$f_4$ & \cellcolor{blue!20}\texttt{1.00E-08$\pm$0.00E+00}$\odot$ & \cellcolor{blue!20}\texttt{1.00E-08$\pm$0.00E+00}$\odot$ & \cellcolor{blue!20}\texttt{1.00E-08$\pm$0.00E+00}$\odot$ & \cellcolor{blue!20}\texttt{1.00E-08$\pm$0.00E+00}$\odot$ & \cellcolor{blue!20}\texttt{1.00E-08$\pm$0.00E+00}\\
$f_5$ & \texttt{6.85E-02$\pm$2.21E-02}$\ominus$ & \cellcolor{blue!20}\texttt{1.39E-03$\pm$5.68E-04}$\oplus$ & \texttt{8.23E-03$\pm$3.22E-03}$\ominus$ & \texttt{1.77E-03$\pm$5.19E-04}$\oplus$ & \texttt{3.18E-03$\pm$9.99E-04}\\
$f_6$ & \texttt{5.90E+01$\pm$1.56E+01}$\ominus$ & \texttt{7.80E-02$\pm$2.70E-01}$\ominus$ & \texttt{1.81E+00$\pm$1.59E+00}$\ominus$ & \texttt{2.36E+01$\pm$6.10E+00}$\ominus$ & \cellcolor{blue!20}\texttt{1.00E-08$\pm$0.00E+00}\\
$f_7$ & \texttt{4.34E+00$\pm$5.77E+00}$\ominus$ & \texttt{8.86E-02$\pm$3.14E-01}$\ominus$ & \texttt{8.57E-01$\pm$8.68E-01}$\ominus$ & \cellcolor{blue!20}\texttt{1.00E-08$\pm$0.00E+00}$\odot$ & \cellcolor{blue!20}\texttt{1.00E-08$\pm$0.00E+00}\\
$f_8$ & \texttt{1.84E-03$\pm$4.59E-03}$\ominus$ & \texttt{4.81E-03$\pm$1.39E-02}$\ominus$ & \texttt{4.40E-02$\pm$9.22E-02}$\ominus$ & \cellcolor{blue!20}\texttt{1.00E-08$\pm$0.00E+00}$\odot$ & \cellcolor{blue!20}\texttt{1.00E-08$\pm$0.00E+00}\\
$f_9$ & \cellcolor{blue!20}\texttt{5.47E-01$\pm$1.39E+00}$\oplus$ & \texttt{3.72E+00$\pm$1.12E+01}$\ominus$ & \texttt{3.22E+01$\pm$2.99E+01}$\ominus$ & \texttt{2.12E+01$\pm$9.22E-01}$\ominus$ & \texttt{1.48E+00$\pm$3.62E+00}\\
$f_{10}$ & \texttt{5.73E-02$\pm$2.31E-01}$\odot$ & \cellcolor{blue!20}\texttt{1.00E-08$\pm$0.00E+00}$\odot$ & \cellcolor{blue!20}\texttt{1.00E-08$\pm$0.00E+00}$\odot$ & \cellcolor{blue!20}\texttt{1.00E-08$\pm$0.00E+00}$\odot$ & \cellcolor{blue!20}\texttt{1.00E-08$\pm$0.00E+00}\\
$f_{11}$ & \texttt{6.46E-04$\pm$2.61E-03}$\odot$ & \cellcolor{blue!20}\texttt{1.00E-08$\pm$0.00E+00}$\odot$ & \texttt{9.48E-02$\pm$3.79E-01}$\ominus$ & \cellcolor{blue!20}\texttt{1.00E-08$\pm$0.00E+00}$\odot$ & \cellcolor{blue!20}\texttt{1.00E-08$\pm$0.00E+00}\\
$f_{12}$ & \texttt{1.37E+01$\pm$3.02E-01}$\ominus$ & \cellcolor{blue!20}\texttt{9.30E-01$\pm$1.10E-01}$\oplus$ & \texttt{1.22E+00$\pm$1.77E-01}$\odot$ & \texttt{1.04E+01$\pm$1.07E+00}$\ominus$ & \texttt{1.27E+00$\pm$1.42E-01}\\
$f_{13}$ & \texttt{4.89E+03$\pm$6.15E+02}$\ominus$ & \texttt{3.25E+02$\pm$1.86E+02}$\ominus$ & \texttt{2.79E+01$\pm$5.60E+01}$\odot$ & \texttt{3.54E+03$\pm$7.98E+02}$\ominus$ & \cellcolor{blue!20}\texttt{1.16E+01$\pm$3.56E+01}\\
$f_{14}$ & \texttt{4.21E+01$\pm$6.16E+01}$\ominus$ & \texttt{6.05E-05$\pm$4.32E-04}$\odot$ & \texttt{2.46E-01$\pm$8.52E-01}$\ominus$ & \texttt{4.01E-01$\pm$2.44E-01}$\ominus$ & \cellcolor{blue!20}\texttt{1.00E-08$\pm$0.00E+00}\\
$f_{15}$ & \texttt{8.02E+01$\pm$1.48E+01}$\ominus$ & \cellcolor{blue!20}\texttt{3.26E+01$\pm$1.57E-01}$\oplus$ & \texttt{3.48E+01$\pm$2.22E+00}$\ominus$ & \texttt{6.03E+01$\pm$7.02E+00}$\ominus$ & \texttt{3.28E+01$\pm$2.30E-01}\\
$f_{16}$ & \texttt{2.32E-03$\pm$5.06E-03}$\ominus$ & \texttt{1.49E-02$\pm$4.05E-02}$\ominus$ & \texttt{3.76E-02$\pm$8.57E-02}$\odot$ & \cellcolor{blue!20}\texttt{1.61E-05$\pm$7.55E-05}$\oplus$ & \texttt{6.76E-04$\pm$6.70E-04}\\
$f_{17}$ & \texttt{8.60E-01$\pm$1.66E+00}$\oplus$ & \cellcolor{blue!20}\texttt{5.47E-01$\pm$1.39E+00}$\oplus$ & \texttt{8.07E+01$\pm$9.08E+01}$\odot$ & \texttt{6.37E+01$\pm$3.32E+01}$\ominus$ & \texttt{3.43E+01$\pm$2.38E+01}\\
$f_{18}$ & \texttt{2.80E-03$\pm$4.84E-03}$\ominus$ & \texttt{8.62E-04$\pm$2.98E-03}$\ominus$ & \texttt{6.94E-01$\pm$2.21E+00}$\ominus$ & \cellcolor{blue!20}\texttt{1.00E-08$\pm$0.00E+00}$\odot$ & \cellcolor{blue!20}\texttt{1.00E-08$\pm$0.00E+00}\\
$f_{19}$ & \texttt{1.38E+01$\pm$2.84E-01}$\ominus$ & \cellcolor{blue!20}\texttt{1.10E+01$\pm$3.24E-01}$\oplus$ & \texttt{1.20E+01$\pm$4.23E-01}$\oplus$ & \texttt{1.33E+01$\pm$6.16E-01}$\ominus$ & \texttt{1.23E+01$\pm$2.42E-01}\\
$f_{20}$ & \texttt{8.54E+01$\pm$1.57E+01}$\oplus$ & \cellcolor{blue!20}\texttt{7.52E+01$\pm$1.09E+01}$\oplus$ & \texttt{9.50E+01$\pm$1.37E+01}$\oplus$ & \texttt{1.17E+02$\pm$6.90E+01}$\ominus$ & \texttt{1.01E+02$\pm$8.78E+00}\\
$f_{21}$ & \texttt{1.76E+03$\pm$3.47E+02}$\ominus$ & \texttt{3.53E+02$\pm$1.93E+02}$\ominus$ & \texttt{6.98E+01$\pm$9.82E+01}$\ominus$ & \texttt{2.41E+02$\pm$1.77E+02}$\ominus$ & \cellcolor{blue!20}\texttt{1.88E+01$\pm$4.34E+01}\\
$f_{22}$ & \texttt{1.55E+02$\pm$4.92E+01}$\ominus$ & \texttt{9.81E+00$\pm$1.38E+01}$\ominus$ & \texttt{1.32E+01$\pm$1.33E+01}$\ominus$ & \texttt{1.11E+01$\pm$2.28E+00}$\ominus$ & \cellcolor{blue!20}\texttt{1.46E+00$\pm$1.69E+00}\\
$f_{23}$ & \texttt{2.88E+01$\pm$6.89E+00}$\ominus$ & \texttt{3.90E+00$\pm$6.57E+00}$\ominus$ & \texttt{4.01E+00$\pm$4.38E+00}$\ominus$ & \cellcolor{blue!20}\texttt{7.43E-01$\pm$1.03E+00}$\oplus$ & \texttt{1.37E+00$\pm$4.32E+00}\\
$f_{24}$ & \texttt{1.30E+03$\pm$2.60E+02}$\ominus$ & \texttt{2.58E+02$\pm$1.52E+02}$\ominus$ & \texttt{3.54E+01$\pm$5.93E+01}$\ominus$ & \texttt{2.38E+01$\pm$2.16E+01}$\ominus$ & \cellcolor{blue!20}\texttt{1.06E+01$\pm$3.98E+01}\\
$f_{25}$ & \texttt{1.18E+03$\pm$3.07E+02}$\ominus$ & \texttt{1.89E+02$\pm$1.35E+02}$\ominus$ & \texttt{4.44E+01$\pm$6.77E+01}$\ominus$ & \texttt{4.06E+01$\pm$4.71E+01}$\ominus$ & \cellcolor{blue!20}\texttt{6.06E+00$\pm$3.35E+00}\\
$\ominus$ & \texttt{16} & \texttt{12} & \texttt{14} & \texttt{13} & \texttt{-}\\
$\oplus$ & \texttt{3} & \texttt{6} & \texttt{2} & \texttt{3} & \texttt{-}\\
$\odot$ & \texttt{6} & \texttt{7} & \texttt{9} & \texttt{9} & \texttt{-}\\
				\hline
        \end{tabular}
    \end{threeparttable}
  \end{center}
\end{sidewaystable}

\begin{figure*}
  \centering \includegraphics[width=\linewidth]{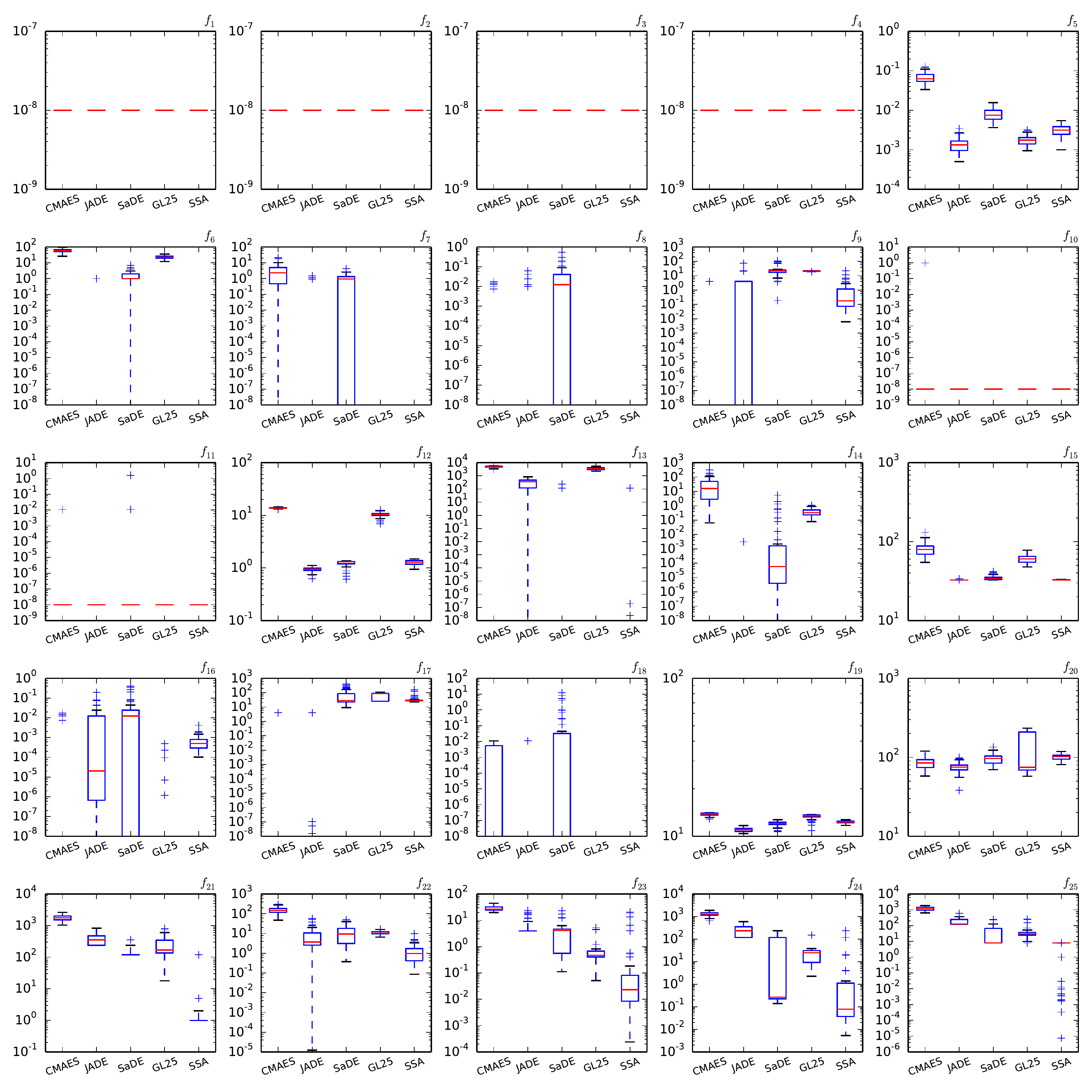}
  \caption{Box plot of raw simulation results.}
  \label{fig:boxplot}
\end{figure*}

We first conduct a series of simulations on the 30-dimension optimization problems using SSA and other state-of-the-art algorithms. The simulation results are plotted in Fig. \ref{fig:boxplot} and Table \ref{tbl:comp30d} reports the means and standard deviations of the optimal fitness values achieved, with the best mean result shaded. Besides, we also perform a series of Wilcoxon rank sum tests on the null hypothesis that SSA performs similarly with other algorithms when solving each benchmark function. The statistical test result at 95\% significance level is presented next to the standard deviation of the corresponding algorithm, where a $\ominus$ indicates that SSA performs significantly better than the tested algorithm on the specified function, a $\oplus$ indicates that SSA performs not as good as the tested algorithm, and a $\odot$ means that the Wilcoxon rank sum test cannot distinguish between the simulation results of SSA and the tested algorithm. The counts of the benchmark functions that fall in these situations are shown at the bottom of the table.

From Table \ref{tbl:comp30d}, we observe that:
\begin{itemize}
\item SSA generally outperforms all compared algorithms in terms of the statistical test. Among all 25 functions, SSA generates better simulation results in 16, 12, 14, and 13 functions compared with CMA-ES, JADE, SaDE, and GL-25, respectively. If we take those functions with similar results, the advantage is more obvious: SSA performs no worse than CMA-ES, JADE, SaDE, and GL-25 in 22, 19, 24, and 22 functions, respectively.
\item In the first group of benchmark functions, all compared algorithms can obtain the global optimum values of $f_1$--$f_4$ in all runs, which means that the final result test cannot reveal the best-performing algorithms. We shall further employ the convergence test to analyse the performance of compared algorithms.
\item The performance of SSA in solving Group II multimodal optimization problems is superior, and it generates 7 best mean results out of the total 10. The numbers for CMA-ES, JADE, SaDE, and GL-25 are 1, 4, 1, and 4, respectively. Besides, the mean results of the three functions where SSA is not the best performing one are still very competitive and comparable to the best results.
\item SSA is not as competitive in solving rotated multimodal functions as it does in unimodal and multimodal functions. However, a careful investigation on the mean results shows that the performance of SSA is still comparable to all compared algorithm. The reason of this phenomenon may be that during the searching process of SSA, no correlation matrix or differential vectors are employed to assist the blind search in the solution space as in CMA-ES and JADE. However, this disadvantage can be overcome by employing these mentioned schemes into SSA or via hybrid algorithms. This is a potential future research direction.
\item SSA is very powerful at solving hybrid functions where different dimensions of the objective functions can be un-related and there is no additional information like correlation matrix available. In Group IV tests, SSA achieved four out of the total five best mean results.
\end{itemize}

\subsection{Scalability Test}

\begin{sidewaystable}[t]
  \caption{Simulation Results for 10-D Problems}
  \label{tbl:comp10d}
  \scriptsize
  \begin{center}
    \begin{threeparttable}
      \begin{tabular}{cccccc}
        \hline
        Function & CMA-ES & JADE & SaDE & GL25 & SSA \\
        & Mean$\pm$Std Dev & Mean$\pm$Std Dev & Mean$\pm$Std Dev & Mean$\pm$Std Dev & Mean$\pm$Std Dev \\
					\hline
$f_1$ & \cellcolor{blue!20}\texttt{1.00E-08$\pm$0.00E+00}$\odot$ & \cellcolor{blue!20}\texttt{1.00E-08$\pm$0.00E+00}$\odot$ & \cellcolor{blue!20}\texttt{1.00E-08$\pm$0.00E+00}$\odot$ & \cellcolor{blue!20}\texttt{1.00E-08$\pm$0.00E+00}$\odot$ & \cellcolor{blue!20}\texttt{1.00E-08$\pm$0.00E+00}\\
$f_2$ & \cellcolor{blue!20}\texttt{1.00E-08$\pm$0.00E+00}$\odot$ & \cellcolor{blue!20}\texttt{1.00E-08$\pm$0.00E+00}$\odot$ & \cellcolor{blue!20}\texttt{1.00E-08$\pm$0.00E+00}$\odot$ & \cellcolor{blue!20}\texttt{1.00E-08$\pm$0.00E+00}$\odot$ & \cellcolor{blue!20}\texttt{1.00E-08$\pm$0.00E+00}\\
$f_3$ & \cellcolor{blue!20}\texttt{1.00E-08$\pm$0.00E+00}$\odot$ & \cellcolor{blue!20}\texttt{1.00E-08$\pm$0.00E+00}$\odot$ & \cellcolor{blue!20}\texttt{1.00E-08$\pm$0.00E+00}$\odot$ & \cellcolor{blue!20}\texttt{1.00E-08$\pm$0.00E+00}$\odot$ & \cellcolor{blue!20}\texttt{1.00E-08$\pm$0.00E+00}\\
$f_4$ & \cellcolor{blue!20}\texttt{1.00E-08$\pm$0.00E+00}$\odot$ & \cellcolor{blue!20}\texttt{1.00E-08$\pm$0.00E+00}$\odot$ & \cellcolor{blue!20}\texttt{1.00E-08$\pm$0.00E+00}$\odot$ & \cellcolor{blue!20}\texttt{1.00E-08$\pm$0.00E+00}$\odot$ & \cellcolor{blue!20}\texttt{1.00E-08$\pm$0.00E+00}\\
$f_5$ & \texttt{2.78E-02$\pm$1.87E-02}$\ominus$ & \cellcolor{blue!20}\texttt{4.58E-04$\pm$2.95E-04}$\oplus$ & \texttt{2.29E-03$\pm$1.98E-03}$\ominus$ & \texttt{4.70E-04$\pm$2.66E-04}$\oplus$ & \texttt{7.50E-04$\pm$3.48E-04}\\
$f_6$ & \texttt{1.60E+01$\pm$7.81E+00}$\ominus$ & \texttt{2.34E+00$\pm$2.48E+00}$\ominus$ & \texttt{2.91E+00$\pm$2.74E+00}$\ominus$ & \texttt{4.61E+00$\pm$1.72E+00}$\ominus$ & \cellcolor{blue!20}\texttt{1.00E-08$\pm$0.00E+00}\\
$f_7$ & \texttt{5.62E-01$\pm$2.90E+00}$\ominus$ & \texttt{4.50E-01$\pm$8.48E-01}$\ominus$ & \texttt{9.14E-01$\pm$1.16E+00}$\ominus$ & \cellcolor{blue!20}\texttt{1.00E-08$\pm$0.00E+00}$\odot$ & \cellcolor{blue!20}\texttt{1.00E-08$\pm$0.00E+00}\\
$f_8$ & \texttt{9.95E-03$\pm$8.75E-03}$\ominus$ & \texttt{4.06E-02$\pm$6.97E-02}$\ominus$ & \texttt{1.15E-01$\pm$1.94E-01}$\ominus$ & \texttt{2.41E-03$\pm$6.32E-03}$\ominus$ & \cellcolor{blue!20}\texttt{1.00E-08$\pm$0.00E+00}\\
$f_9$ & \texttt{3.13E-01$\pm$1.08E+00}$\ominus$ & \texttt{5.47E-01$\pm$1.39E+00}$\ominus$ & \texttt{3.08E+00$\pm$7.04E+00}$\ominus$ & \texttt{2.34E+00$\pm$6.81E-01}$\ominus$ & \cellcolor{blue!20}\texttt{2.60E-01$\pm$7.78E-01}\\
$f_{10}$ & \texttt{3.82E-02$\pm$1.91E-01}$\odot$ & \texttt{4.88E-02$\pm$1.14E-01}$\ominus$ & \texttt{7.32E-02$\pm$3.96E-01}$\ominus$ & \cellcolor{blue!20}\texttt{1.00E-08$\pm$0.00E+00}$\odot$ & \cellcolor{blue!20}\texttt{1.00E-08$\pm$0.00E+00}\\
$f_{11}$ & \texttt{1.72E-03$\pm$4.04E-03}$\ominus$ & \texttt{3.42E-03$\pm$1.50E-02}$\ominus$ & \texttt{1.26E-01$\pm$5.57E-01}$\ominus$ & \cellcolor{blue!20}\texttt{1.00E-08$\pm$0.00E+00}$\odot$ & \cellcolor{blue!20}\texttt{1.00E-08$\pm$0.00E+00}\\
$f_{12}$ & \texttt{4.24E+00$\pm$2.78E-01}$\ominus$ & \texttt{2.83E-01$\pm$2.46E-01}$\odot$ & \texttt{4.18E-01$\pm$1.11E-01}$\ominus$ & \texttt{1.16E+00$\pm$7.46E-01}$\ominus$ & \cellcolor{blue!20}\texttt{2.20E-01$\pm$8.60E-02}\\
$f_{13}$ & \texttt{1.61E+03$\pm$3.69E+02}$\ominus$ & \texttt{3.19E+02$\pm$1.81E+02}$\ominus$ & \texttt{1.15E+02$\pm$1.18E+02}$\ominus$ & \texttt{2.67E+02$\pm$2.21E+02}$\ominus$ & \cellcolor{blue!20}\texttt{2.32E+00$\pm$1.66E+01}\\
$f_{14}$ & \texttt{1.36E+01$\pm$2.91E+01}$\ominus$ & \texttt{3.88E-01$\pm$2.51E+00}$\ominus$ & \texttt{1.44E+00$\pm$3.49E+00}$\ominus$ & \texttt{3.17E-03$\pm$5.31E-03}$\ominus$ & \cellcolor{blue!20}\texttt{1.00E-08$\pm$0.00E+00}\\
$f_{15}$ & \texttt{1.36E+01$\pm$7.35E+00}$\ominus$ & \texttt{6.38E+00$\pm$6.57E+00}$\odot$ & \texttt{1.14E+01$\pm$6.65E+00}$\ominus$ & \texttt{1.19E+01$\pm$3.78E+00}$\ominus$ & \cellcolor{blue!20}\texttt{4.04E+00$\pm$4.96E+00}\\
$f_{16}$ & \texttt{9.85E-03$\pm$9.45E-03}$\odot$ & \texttt{2.63E-02$\pm$3.09E-02}$\ominus$ & \texttt{1.17E-01$\pm$1.06E-01}$\ominus$ & \cellcolor{blue!20}\texttt{2.82E-03$\pm$4.99E-03}$\oplus$ & \texttt{4.35E-03$\pm$5.12E-03}\\
$f_{17}$ & \cellcolor{blue!20}\texttt{4.69E-01$\pm$1.30E+00}$\oplus$ & \texttt{9.38E-01$\pm$1.71E+00}$\oplus$ & \texttt{1.75E+01$\pm$3.36E+01}$\ominus$ & \texttt{4.91E+00$\pm$6.39E-01}$\oplus$ & \texttt{1.49E+01$\pm$3.74E+01}\\
$f_{18}$ & \texttt{1.70E-03$\pm$4.51E-03}$\ominus$ & \texttt{1.29E-03$\pm$3.58E-03}$\ominus$ & \texttt{1.87E-02$\pm$3.82E-02}$\ominus$ & \cellcolor{blue!20}\texttt{1.00E-08$\pm$0.00E+00}$\odot$ & \cellcolor{blue!20}\texttt{1.00E-08$\pm$0.00E+00}\\
$f_{19}$ & \texttt{4.21E+00$\pm$2.21E-01}$\ominus$ & \cellcolor{blue!20}\texttt{2.76E+00$\pm$5.53E-01}$\oplus$ & \texttt{3.26E+00$\pm$3.83E-01}$\odot$ & \texttt{3.09E+00$\pm$4.48E-01}$\odot$ & \texttt{3.13E+00$\pm$2.32E-01}\\
$f_{20}$ & \texttt{1.52E+01$\pm$6.98E+00}$\ominus$ & \texttt{7.67E+00$\pm$5.05E+00}$\odot$ & \texttt{1.54E+01$\pm$8.07E+00}$\ominus$ & \cellcolor{blue!20}\texttt{6.02E+00$\pm$3.21E+00}$\odot$ & \texttt{6.41E+00$\pm$2.46E+00}\\
$f_{21}$ & \texttt{7.07E+02$\pm$2.56E+02}$\ominus$ & \texttt{2.18E+02$\pm$1.52E+02}$\ominus$ & \texttt{6.46E+01$\pm$9.33E+01}$\ominus$ & \cellcolor{blue!20}\texttt{1.74E+00$\pm$3.21E+00}$\oplus$ & \texttt{2.34E+00$\pm$1.66E+01}\\
$f_{22}$ & \texttt{4.56E+01$\pm$2.88E+01}$\ominus$ & \texttt{1.06E+01$\pm$1.74E+01}$\ominus$ & \texttt{5.71E+00$\pm$7.26E+00}$\ominus$ & \texttt{4.02E-01$\pm$4.75E-01}$\ominus$ & \cellcolor{blue!20}\texttt{6.95E-02$\pm$1.42E-01}\\
$f_{23}$ & \texttt{2.13E+01$\pm$4.43E+00}$\ominus$ & \texttt{3.44E+00$\pm$7.53E+00}$\ominus$ & \texttt{3.31E+00$\pm$7.02E+00}$\odot$ & \cellcolor{blue!20}\texttt{7.76E-03$\pm$4.06E-02}$\oplus$ & \texttt{4.70E-01$\pm$2.85E+00}\\
$f_{24}$ & \texttt{4.07E+02$\pm$1.66E+02}$\ominus$ & \texttt{1.03E+02$\pm$8.79E+01}$\ominus$ & \texttt{3.05E+01$\pm$6.38E+01}$\ominus$ & \cellcolor{blue!20}\texttt{1.15E+00$\pm$1.34E+00}$\oplus$ & \texttt{4.72E+00$\pm$3.32E+01}\\
$f_{25}$ & \texttt{3.22E+02$\pm$1.63E+02}$\ominus$ & \texttt{1.00E+02$\pm$1.05E+02}$\ominus$ & \texttt{4.88E+01$\pm$7.23E+01}$\ominus$ & \cellcolor{blue!20}\texttt{1.95E-01$\pm$4.46E-01}$\oplus$ & \texttt{2.32E+00$\pm$1.66E+01}\\
$\ominus$ & \texttt{18} & \texttt{15} & \texttt{19} & \texttt{8} & \texttt{-}\\
$\oplus$ & \texttt{1} & \texttt{3} & \texttt{0} & \texttt{7} & \texttt{-}\\
$\odot$ & \texttt{6} & \texttt{7} & \texttt{6} & \texttt{10} & \texttt{-}\\
				\hline
        \end{tabular}
    \end{threeparttable}
  \end{center}
\end{sidewaystable}

\begin{sidewaystable}[t]
  \caption{Simulation Results for 50-D Problems}
  \label{tbl:comp50d}
  \scriptsize
  \begin{center}
    \begin{threeparttable}
      \begin{tabular}{cccccc}
        \hline
        Function & CMA-ES & JADE & SaDE & GL25 & SSA \\
        & Mean$\pm$Std Dev & Mean$\pm$Std Dev & Mean$\pm$Std Dev & Mean$\pm$Std Dev & Mean$\pm$Std Dev \\
					\hline
$f_1$ & \cellcolor{blue!20}\texttt{1.00E-08$\pm$0.00E+00}$\odot$ & \cellcolor{blue!20}\texttt{1.00E-08$\pm$0.00E+00}$\odot$ & \cellcolor{blue!20}\texttt{1.00E-08$\pm$0.00E+00}$\odot$ & \cellcolor{blue!20}\texttt{1.00E-08$\pm$0.00E+00}$\odot$ & \cellcolor{blue!20}\texttt{1.00E-08$\pm$0.00E+00}\\
$f_2$ & \texttt{1.32E-08$\pm$1.69E-08}$\odot$ & \cellcolor{blue!20}\texttt{1.00E-08$\pm$0.00E+00}$\odot$ & \cellcolor{blue!20}\texttt{1.00E-08$\pm$0.00E+00}$\odot$ & \cellcolor{blue!20}\texttt{1.00E-08$\pm$0.00E+00}$\odot$ & \cellcolor{blue!20}\texttt{1.00E-08$\pm$0.00E+00}\\
$f_3$ & \cellcolor{blue!20}\texttt{1.00E-08$\pm$0.00E+00}$\odot$ & \cellcolor{blue!20}\texttt{1.00E-08$\pm$0.00E+00}$\odot$ & \cellcolor{blue!20}\texttt{1.00E-08$\pm$0.00E+00}$\odot$ & \cellcolor{blue!20}\texttt{1.00E-08$\pm$0.00E+00}$\odot$ & \cellcolor{blue!20}\texttt{1.00E-08$\pm$0.00E+00}\\
$f_4$ & \cellcolor{blue!20}\texttt{1.00E-08$\pm$0.00E+00}$\odot$ & \cellcolor{blue!20}\texttt{1.00E-08$\pm$0.00E+00}$\odot$ & \cellcolor{blue!20}\texttt{1.00E-08$\pm$0.00E+00}$\odot$ & \cellcolor{blue!20}\texttt{1.00E-08$\pm$0.00E+00}$\odot$ & \cellcolor{blue!20}\texttt{1.00E-08$\pm$0.00E+00}\\
$f_5$ & \texttt{7.88E+00$\pm$5.55E+01}$\ominus$ & \cellcolor{blue!20}\texttt{2.25E-03$\pm$8.17E-04}$\oplus$ & \texttt{1.65E-02$\pm$6.08E-03}$\ominus$ & \texttt{4.22E-03$\pm$1.45E-03}$\oplus$ & \texttt{6.26E-03$\pm$1.32E-03}\\
$f_6$ & \texttt{1.86E+02$\pm$2.76E+02}$\ominus$ & \cellcolor{blue!20}\texttt{1.00E-08$\pm$0.00E+00}$\odot$ & \texttt{1.31E+00$\pm$1.10E+00}$\ominus$ & \texttt{4.98E+01$\pm$1.18E+01}$\ominus$ & \cellcolor{blue!20}\texttt{1.00E-08$\pm$0.00E+00}\\
$f_7$ & \texttt{1.38E+01$\pm$8.17E+00}$\ominus$ & \texttt{1.60E-01$\pm$3.78E-01}$\ominus$ & \texttt{9.60E-01$\pm$6.57E-01}$\ominus$ & \cellcolor{blue!20}\texttt{1.00E-08$\pm$0.00E+00}$\odot$ & \cellcolor{blue!20}\texttt{1.00E-08$\pm$0.00E+00}\\
$f_8$ & \texttt{5.80E-04$\pm$2.01E-03}$\ominus$ & \texttt{8.13E-03$\pm$1.70E-02}$\ominus$ & \texttt{3.89E-02$\pm$6.64E-02}$\ominus$ & \cellcolor{blue!20}\texttt{1.00E-08$\pm$0.00E+00}$\odot$ & \cellcolor{blue!20}\texttt{1.00E-08$\pm$0.00E+00}\\
$f_9$ & \cellcolor{blue!20}\texttt{1.56E-01$\pm$7.82E-01}$\oplus$ & \texttt{1.02E+00$\pm$1.75E+00}$\oplus$ & \texttt{7.36E+01$\pm$2.81E+01}$\ominus$ & \texttt{4.13E+01$\pm$1.25E+00}$\ominus$ & \texttt{4.34E+00$\pm$9.45E+00}\\
$f_{10}$ & \texttt{3.35E+00$\pm$1.22E+01}$\ominus$ & \texttt{2.44E-03$\pm$1.22E-02}$\odot$ & \texttt{3.66E-03$\pm$1.93E-02}$\odot$ & \cellcolor{blue!20}\texttt{1.00E-08$\pm$0.00E+00}$\odot$ & \cellcolor{blue!20}\texttt{1.00E-08$\pm$0.00E+00}\\
$f_{11}$ & \texttt{8.62E-04$\pm$2.98E-03}$\ominus$ & \texttt{2.15E-04$\pm$1.54E-03}$\odot$ & \texttt{2.82E-01$\pm$9.77E-01}$\ominus$ & \texttt{6.57E-02$\pm$1.09E-01}$\ominus$ & \cellcolor{blue!20}\texttt{1.00E-08$\pm$0.00E+00}\\
$f_{12}$ & \texttt{2.39E+01$\pm$8.29E-01}$\ominus$ & \cellcolor{blue!20}\texttt{1.73E+00$\pm$1.96E-01}$\oplus$ & \texttt{4.84E+00$\pm$1.33E+00}$\oplus$ & \texttt{1.91E+01$\pm$1.53E+00}$\ominus$ & \texttt{6.30E+00$\pm$6.35E-01}\\
$f_{13}$ & \texttt{8.45E+03$\pm$7.98E+02}$\ominus$ & \texttt{1.14E+02$\pm$1.18E+02}$\ominus$ & \texttt{4.64E+00$\pm$2.32E+01}$\odot$ & \texttt{7.39E+03$\pm$8.76E+02}$\ominus$ & \cellcolor{blue!20}\texttt{1.00E-08$\pm$0.00E+00}\\
$f_{14}$ & \texttt{1.90E+02$\pm$1.40E+02}$\ominus$ & \cellcolor{blue!20}\texttt{1.00E-08$\pm$0.00E+00}$\odot$ & \texttt{2.96E-01$\pm$8.43E-01}$\ominus$ & \texttt{8.74E+00$\pm$3.56E+00}$\ominus$ & \cellcolor{blue!20}\texttt{1.00E-08$\pm$0.00E+00}\\
$f_{15}$ & \texttt{1.53E+02$\pm$2.51E+01}$\ominus$ & \cellcolor{blue!20}\texttt{5.00E+01$\pm$6.38E-14}$\oplus$ & \texttt{5.23E+01$\pm$1.72E+00}$\oplus$ & \texttt{1.07E+02$\pm$1.11E+01}$\ominus$ & \texttt{5.71E+01$\pm$4.62E+00}\\
$f_{16}$ & \cellcolor{blue!20}\texttt{5.80E-04$\pm$2.39E-03}$\oplus$ & \texttt{2.20E-02$\pm$2.63E-02}$\odot$ & \texttt{2.49E-02$\pm$2.81E-02}$\odot$ & \texttt{1.42E-02$\pm$1.85E-02}$\odot$ & \texttt{1.31E-02$\pm$7.03E-03}\\
$f_{17}$ & \cellcolor{blue!20}\texttt{5.47E-01$\pm$1.39E+00}$\oplus$ & \texttt{9.46E-01$\pm$1.71E+00}$\oplus$ & \texttt{6.22E+01$\pm$3.07E+01}$\odot$ & \texttt{4.70E+01$\pm$1.02E+00}$\ominus$ & \texttt{4.46E+01$\pm$2.63E+00}\\
$f_{18}$ & \texttt{1.08E-03$\pm$3.30E-03}$\ominus$ & \texttt{1.29E-03$\pm$3.58E-03}$\ominus$ & \texttt{5.45E+00$\pm$1.60E+01}$\ominus$ & \texttt{1.31E-02$\pm$2.86E-02}$\ominus$ & \cellcolor{blue!20}\texttt{1.00E-08$\pm$0.00E+00}\\
$f_{19}$ & \texttt{2.41E+01$\pm$7.83E-01}$\ominus$ & \cellcolor{blue!20}\texttt{1.98E+01$\pm$3.91E-01}$\oplus$ & \texttt{2.17E+01$\pm$2.58E-01}$\oplus$ & \texttt{2.34E+01$\pm$2.50E-01}$\ominus$ & \texttt{2.19E+01$\pm$2.19E-01}\\
$f_{20}$ & \texttt{1.59E+02$\pm$1.92E+01}$\oplus$ & \cellcolor{blue!20}\texttt{1.19E+02$\pm$1.03E+01}$\oplus$ & \texttt{1.82E+02$\pm$2.54E+01}$\oplus$ & \texttt{1.99E+02$\pm$1.25E+02}$\oplus$ & \texttt{2.49E+02$\pm$1.62E+01}\\
$f_{21}$ & \texttt{3.29E+03$\pm$5.11E+02}$\ominus$ & \texttt{5.74E+02$\pm$2.50E+02}$\ominus$ & \texttt{3.96E+01$\pm$8.09E+01}$\ominus$ & \texttt{1.42E+03$\pm$4.17E+02}$\ominus$ & \cellcolor{blue!20}\texttt{1.89E+01$\pm$4.34E+01}\\
$f_{22}$ & \texttt{2.33E+02$\pm$6.38E+01}$\ominus$ & \texttt{2.33E+01$\pm$1.84E+01}$\ominus$ & \texttt{3.24E+01$\pm$1.80E+01}$\ominus$ & \texttt{2.40E+01$\pm$4.10E+00}$\ominus$ & \cellcolor{blue!20}\texttt{3.75E+00$\pm$3.37E+00}\\
$f_{23}$ & \texttt{2.87E+01$\pm$8.78E+00}$\ominus$ & \texttt{1.04E+01$\pm$1.12E+01}$\ominus$ & \texttt{1.20E+01$\pm$2.07E+01}$\ominus$ & \texttt{4.68E+00$\pm$7.69E-01}$\ominus$ & \cellcolor{blue!20}\texttt{5.71E-01$\pm$1.56E+00}\\
$f_{24}$ & \texttt{2.16E+03$\pm$4.29E+02}$\ominus$ & \texttt{3.36E+02$\pm$1.88E+02}$\ominus$ & \texttt{5.93E+01$\pm$8.32E+01}$\odot$ & \texttt{2.10E+02$\pm$1.62E+02}$\ominus$ & \cellcolor{blue!20}\texttt{1.42E+01$\pm$2.82E+01}\\
$f_{25}$ & \texttt{1.95E+03$\pm$3.67E+02}$\ominus$ & \texttt{3.11E+02$\pm$1.80E+02}$\ominus$ & \texttt{3.02E+01$\pm$5.72E+01}$\ominus$ & \texttt{1.94E+02$\pm$1.56E+02}$\ominus$ & \cellcolor{blue!20}\texttt{6.97E+00$\pm$2.81E+01}\\
$\ominus$ & \texttt{17} & \texttt{9} & \texttt{12} & \texttt{15} & \texttt{-}\\
$\oplus$ & \texttt{4} & \texttt{7} & \texttt{4} & \texttt{2} & \texttt{-}\\
$\odot$ & \texttt{4} & \texttt{9} & \texttt{9} & \texttt{8} & \texttt{-}\\
				\hline
        \end{tabular}
    \end{threeparttable}
  \end{center}
\end{sidewaystable}

In addition to the 30-dimension benchmark function tests, we also performed a series of simulations on both 10- and 50-dimension benchmarks to test the scalability of SSA. To make a thorough comparison, we also employed the compared algorithms in this test. The simulation results are presented in Tables \ref{tbl:comp10d} and \ref{tbl:comp50d}, using the same format and symbols as in Table \ref{tbl:comp30d}. From the results we have the following observations:
\begin{itemize}
\item The advantage of SSA over compared algorithms are confirmed. SSA achieved all the best mean results in 10-D Group II tests and 50-D Group IV tests, and have satisfactory performance compared with other algorithms in all other groups.
\item This advantage is also supported by the statistical test. SSA can generate better results than CMA-ES, JADE, SaDE, and GL-25 in 18, 15, 19, and 8 10-D functions, respectively. The corresponding numbers for 50-D functions are 17, 9, 12, and 15, respectively. The statistical results will favor SSA more if we also take those tests that have similar performance into account.
\item From the simulation results we can see GL-25 performs very well in 10-D hybrid functions, JADE performs very well in 50-D multimodal functions, when compared with the remaining three algorithms. However, SSA can always outperform them in these tests, which indicate the superior scalability of SSA.
\end{itemize}

\subsection{Convergence Test}

\begin{figure*}
  \centering \includegraphics[width=\linewidth]{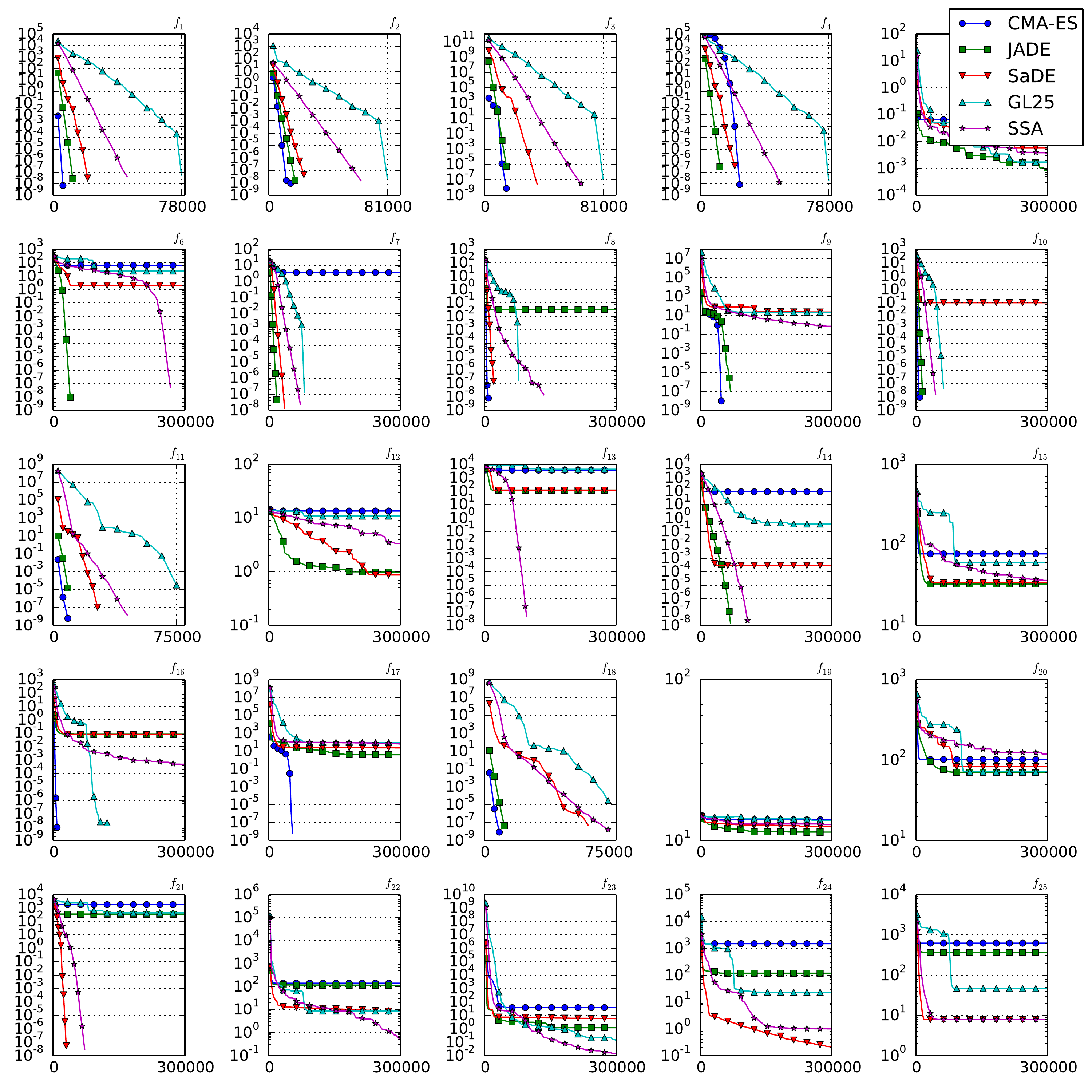}
  \caption{Median convergence test results.}
  \label{fig:convergence}
\end{figure*}

As stated in Section \ref{sub:compare}, the final result comparison cannot completely describe the searching performance of an algorithm. So we further conduct a convergence test on the five compared algorithms on each 30-D benchmark function. We employ the raw simulation data generated in Section \ref{sub:compare}. As each function is tested for 51 runs for each algorithm, we select the convergence data of the run which generates the median final result. The convergence data of the five compared algorithms are plotted in Fig. \ref{fig:convergence}. The x-axis is the function evaluations consumed, and the y-axis is the best-so-far fitness values found. The convergence plots lead to the following observations:
\begin{itemize}
\item The convergence speed of SSA in solving unimodal optimization problems is not as fast as CMA-ES, JADE, and SaDE. This is because SSA performs exploitation and exploration simultaneously during the random walk process, and the former contributes to a fast convergence speed but the latter obstructs the population from moving into a small region in the search space. However, the degree of exploration in the random walk process can be effectively controlled by the algorithm parameters as revealed in Section \ref{sub:param}. So it is highly possible that SSA can also achieve a comparable convergence speed with other algorithms if we adopt a set of suitable parameters designed for solving unimodal optimization problems.
\item When solving multi-modal optimization problems, SSA generally converges as fast as or even faster than the compared algorithms. This phenomenon can be clearly observed in the Group II tests, where SSA mostly generates a similar convergence curve with others.
\item The advantage of combining exploitation and exploration in one searching process is revealed in the convergence plot. Take $f_6$ as an example, almost all algorithms (except JADE) are trapped in local optima shortly after the start of searching. However, SSA managed to jump out of the optimum and successfully found the global optimum by the end of searching. The manifestation of this searching characteristic is that the algorithm convergences relatively slow at the beginning, but then very fast as the search continues. This phenomenon can also be observed in some other instances, e.g. $f_{13}$, $f_{22}$, and $f_{24}$.
\end{itemize}

\subsection{Reliability Test}

\begin{figure*}
  \centering \includegraphics[width=\linewidth]{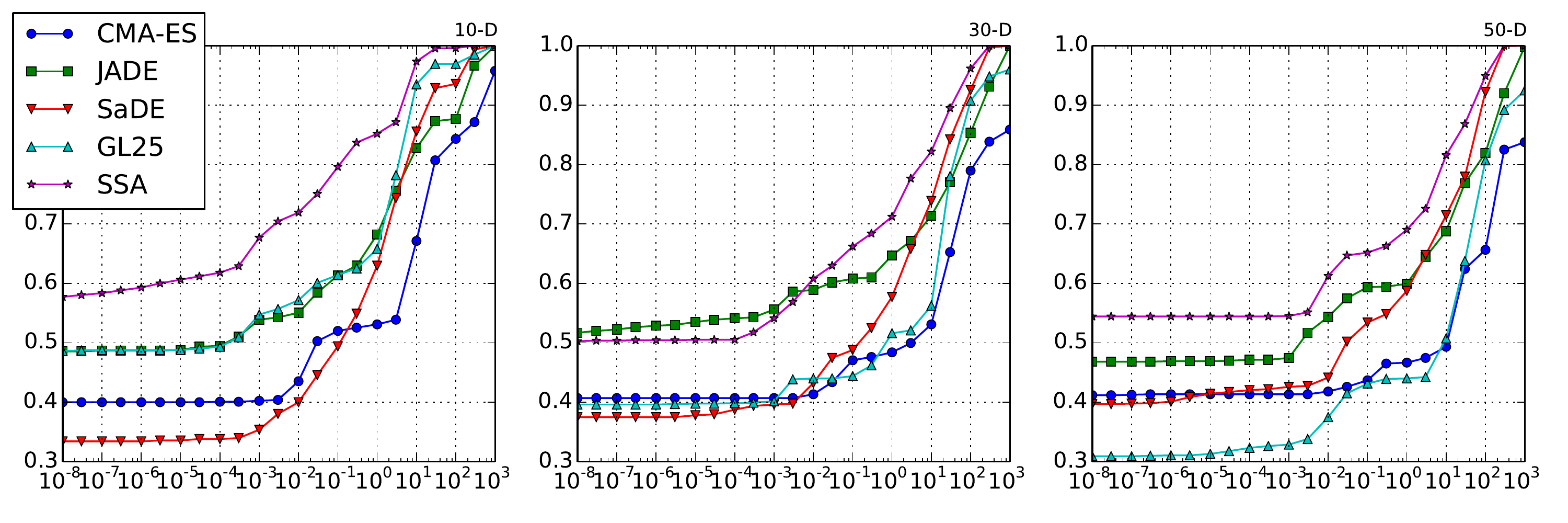}
  \caption{Empirical cumulative distribution of success rate test results.}
  \label{fig:success}
\end{figure*}

Another test which can examine the reliability of stochastic algorithms is the success rate test and this test has been adopted in many previous work, e.g. \cite{QinHuangSuganthan2009DifferentialEvolutionAlgorithm}\cite{ZhanZhangLiChung2009AdaptiveParticleSwarm}. In this part, we aim at further comparing and visualizing the performance of all compared algorithms on all tested benchmark functions by plotting the empirical cumulative distribution of success rates. The plots are presented in Fig. \ref{fig:success}, where the x-axis is different success thresholds, and the y-axis is the overall success rate. A simulation run is considered successful if and only if the best-found fitness value is smaller than or equal to the ``successful threshold''. The overall success rate is calculated by the number of successful runs under a specific success threshold divided by the total number of runs. Thus a larger overall success rate implies a more reliable algorithm. By comparing the distribution curves of different algorithms, we can have a general view on which algorithm is reliable at solving general optimization problems.

From the plots we have the following observations:
\begin{itemize}
\item SSA is generally more reliable than the compared algorithms. The advantage is very significant in 10-D and 50-D benchmark functions. While the reliability of SSA is very similar to JADE in small success threshold regions, SSA regains the lead in large threshold regions.
\item The plot also demonstrates the convergence characteristic of SSA. SSA is able to find the global optimum whenever the algorithm is able to locate a relatively small region near it. This conclusion is drawn based on the observation that almost all runs that obtained a fitness value smaller than $10^{-4}$ in 30-D functions, and $10^{-3}$ in 50-D functions are able to converge to the global optimum point at $10^{-8}$.
\end{itemize}

\subsection{Computational Complexity}

Besides the previous tests, the computational complexity is also a major factor for evaluating the efficiency of an evolutionary computation algorithm. In this paper, we employ the method stated in \cite{LiangQuSuganthanHernandez-Diaz2013ProblemDefinitionsand} and \cite{LiangQuSuganthan2014ProblemDefinitionsand} to analyze the computational complexity of the compared algorithms. We use $f_6$ in Table \ref{tbl:benchmark}, which is the major component of the testing methodology suggested by \cite{LiangQuSuganthan2014ProblemDefinitionsand}, as the benchmark evaluation function, and the complexity analysis result is as follows. The complexity values of CMA-ES, JADE, SaDE, GL-25, and SSA are 40.47, 34.69, 75.72, 63.51, and 44.18, respectively. A smaller complexity value means that the algorithm is less complex, which leads to a relatively faster speed in execution under the same condition. From the results we can see that although SSA is slightly more complicated than JADE and CMA-ES, their complexities are comparable. In addition, these three algorithms are significantly less computationally complex than SaDE and GL-25.

\subsection{Discussion}

As stated by the No-Free-Lunch (NFL) Theorem \cite{WolpertMacready1997NoFreeLunch}, all meta-heuristics that search for extrema shall perform exactly the same when all possible objective functions are evaluated and averaged. It is further elaborated that it is theoretically impossible to have a best general-purpose universal optimization technique \cite{HoPepyne2002Simpleexplanationno}. Superior performing algorithms are available if particular classes of problems are considered \cite{HoPepyne2002Simpleexplanationno}, or general but real-world ones \cite{Garcia-MartinezRodriguezLozano2012Arbitraryfunctionoptimisation}.

However, the total number of possible problems are so huge that there is still much room to develop new algorithms. Though existing meta-heuristics have great success in solving many optimization problems, it is always worthwhile to develop new searching methodologies with superior performance in particular classes of problems \cite{LamLi2010ChemicalReactionInspired}. This is the motivation for us to propose SSA for solving global numerical optimization problems.

\section{Conclusion}\label{sec:conclusion}

In this paper we proposed a novel social spider algorithm to solve global optimization problems. This algorithm is based on the foraging behavior of social spiders and the information-sharing foraging strategy. SSA is conceptually simple and relatively easy to implement. SSA can tackle a wide range of different continuous optimization problems and has the potential to be employed to solve real-world problems.

In order to evaluate the performance of SSA, we adopted a set of 25 benchmark functions which cover a large variety of different optimization problem types. We compared SSA with the state-of-the-art optimization algorithms, namely, CMA-ES, JADE, SaDE, and GL-25. These algorithms have been employed to solve a large set of different benchmark optimization functions and real-world problems, and demonstrated outstanding performance. The results show that the performance of SSA is outstanding compared with the above listed algorithms in all three different groups of functions. This conclusion was supported by both the simulation results and the statistics of the simulation data.

Future research on SSA can be divided into three categories: scheme research, algorithm research, and real-world application. The random walk scheme in the current SSA may be further improved using advanced optimization techniques and hybrid algorithms with deterministic heuristics or local search algorithms. New schemes can also be applied in the searching process of SSA for performance improvement. In terms of algorithm research, SSA has the potential to be applied to solve combinatorial problems. We note that some other swarm intelligence algorithms like PSO and ABC originally designed to solve continuous optimization problems have been successfully modified to solve combinatorial problem \cite{KennedyEberhart1997discretebinaryversion}\cite{PanTasgetirenSuganthanChua2011discreteartificialbee}. Although SSA only has three parameters besides the population size, it is still very interesting to develop adaptive or self-adaptive schemes for SSA to control the parameters and reduce the effort in tuning parameters. Last but not least, it would be interesting to identify real-world applications which can be addressed using SSA effectively and efficiently.

\section*{Acknowledgment}
The authors would like to thank the anonymous reviewer for useful, constructive comments.

\section*{References}

\bibliographystyle{elsarticle-num}
\bibliography{IEEEabrv,../../../../bib/publications}

\section*{Biography}

\noindent{\bf James J.Q. Yu} received the B.Eng. degree in Electrical and Electronic Engineering from the University of Hong Kong, Pokfulam, Hong Kong, in 2011. He is now a Ph.D. candidate at the Department of Electrical and Electronic Engineering of the University of Hong Kong. His current research interests include optimization algorithm design and analysis, evolutionary computation and its application, data mining, wireless communications, and power system.

\noindent{\bf Victor O.K. Li} received SB, SM, EE and ScD degrees in Electrical Engineering and Computer Science from MIT in 1977, 1979, 1980, and 1981, respectively. He is Chair Professor of Information Engineering and Head of the Department of Electrical and Electronic Engineering at the University of Hong Kong (HKU). He also served as Associate Dean of Engineering, and Managing Director of Versitech Ltd., the technology transfer and commercial arm of HKU, and on the board of China.com Ltd. He is now serving on the boards of Sunevision Holdings Ltd. and Anxin-China Holdings Ltd., listed on the Hong Kong Stock Exchange. Previously, he was Professor of Electrical Engineering at the University of Southern California (USC), Los Angeles, California, USA, and Director of the USC Communication Sciences Institute. Sought by government, industry, and academic organizations, he has lectured and consulted extensively around the world. He has received numerous awards, including the PRC Ministry of Education Changjiang Chair Professorship at Tsinghua University, the UK Royal Academy of Engineering Senior Visiting Fellowship in Communications, the Croucher Foundation Senior Research Fellowship, and the Order of the Bronze Bauhinia Star, Government of the Hong Kong Special Administrative Region, China. He is a Registered Professional Engineer and a Fellow of the Hong Kong Academy of Engineering Sciences, the IEEE, the IAE, and the HKIE.

\end{document}